\documentclass[11pt,a4paper]{article}
\usepackage{times,latexsym}
\usepackage{url}
\usepackage[T1]{fontenc}

\usepackage{graphicx}
\usepackage{subcaption}

\usepackage{caption}

\usepackage{enumitem}
\usepackage{amsmath}
\usepackage{hyperref}
\usepackage{multirow}

\usepackage{colortbl}
\usepackage{xcolor}
\usepackage{booktabs}

\usepackage{tipa}
\usepackage{makecell}

\usepackage{soul}

\usepackage{siunitx}

\usepackage{float}
\usepackage{twemojis}

\usepackage[acceptedWithA]{tacl2021v1}

\usepackage{xspace,mfirstuc,tabulary}

\newcommand{\finding}{\scalebox{1.5}{\twemoji{1f4cc}}}

\newif\iftaclinstructions
\taclinstructionsfalse 
\iftaclinstructions
\renewcommand{\confidential}{}
\renewcommand{\anonsubtext}{(No author info supplied here, for consistency with
TACL-submission anonymization requirements)}
\newcommand{\instr}
\fi

\iftaclpubformat 

\else

\fi

\title{SPES: Spectrogram Perturbation \\ for Explainable Speech-to-Text Generation}

\author{
Dennis Fucci*$^\diamond$, Marco Gaido$^\diamond$, Beatrice Savoldi$^\diamond$, \\
\textbf{Matteo Negri}$^\diamond$, \textbf{Mauro Cettolo}$^\diamond$, \textbf{Luisa Bentivogli}$^\diamond$
  \\
 *University of Trento, Italy
 \\
 $^\diamond$Fondazione Bruno Kessler, Italy
 \\
   \texttt{\{dfucci,mgaido,bsavoldi,negri,cettolo,bentivo\}@fbk.eu}
   }

\date{}

\begin{document}
\maketitle
\begin{abstract}
Spurred by the demand for interpretable models, research on eXplainable AI for language technologies has experienced significant growth, with feature attribution methods emerging as a cornerstone of this progress. 
While prior work in NLP explored such methods for \textit{classification}
tasks and \textit{textual} applications, explainability
intersecting 
\textit{generation} and \textit{speech} is lagging, with existing techniques failing to account for the autoregressive nature of state-of-the-art models and to provide fine-grained, phonetically meaningful explanations.
We address this gap by introducing
\textit{\textbf{S}pectrogram \textbf{P}erturbation for \textbf{E}xplainable \textbf{S}peech-to-text Generation} (\textbf{SPES}), a feature attribution technique applicable to sequence generation tasks with autoregressive models. SPES provides explanations for each predicted token based on both the input spectrogram and the previously generated tokens. Extensive evaluation on speech recognition and translation demonstrates that SPES generates explanations that are faithful and plausible to humans.
\end{abstract}

\section{Introduction}

The recent advances of Artificial Intelligence (AI) and the emergence of foundation models \citep{Bommasani2021FoundationModels} have come together with concerns about their risks and impact \citep{weidinger2021ethicalsocialrisksharm}, as well as calls for a better understanding of their inner workings \citep{arrieta-etal-2020,eiras2024risks}. 
This need has been reinforced by the demands for transparency by legal frameworks like the EU AI Act and the US National AI Initiative Act \citep{panigutti-etal-2023, huw-etal-2024}.
In response, the field of eXplainable AI (XAI) has emerged prominently, with the goal of elucidating the reasoning behind system 
predictions \citep{doshivelez-kim-2017, carvalho-etal-2019, vilone-longo-2021, pradhan-etal-2022}.
Among the various active XAI approaches \citep{ferrando-etal-2024}, this paper centers on \textit{feature attribution} methods, which aim to identify and quantify the importance of each input feature in determining a model's final output.

Originally developed for image and text classification \citep{danilevsky-etal-2020-survey,kamakshi-krishnan-2023}, feature attribution methods have been extended to other modalities---like speech \citep{becker-etal-2019, pastor-etal-2024-explaining}---and to text generation \citep[][et al.]{sarti-etal-2023-inseq, zhao-etal-2024b}.
Still, XAI in the domain of speech-to-text (S2T) generative tasks is limited, with only a few preliminary works focusing on automatic speech recognition (ASR) \citep{mandel16_bubble, trinh18_interspeech, trinh21_bubble, kavaki20_interspeech, markert-etal-2021, wu-etal-2023-explanations, wu-etal-2024-trust}. 
This contrasts with the pivotal role of spoken language---arguably the most natural form of human interaction \citep{munteanu-etal-2013-hci}---and consequently with the importance of S2T technologies, which are now fostered by foundation models that transcribe and translate at an unprecedented scale \citep{latif2023sparks}.

Moreveor, most existing feature attribution methods in S2T \cite{mandel16_bubble, trinh21_bubble, markert-etal-2021, wu-etal-2023-explanations, wu-etal-2024-trust} are not applied to autoregressive models, which predict each token iteratively, relying on both speech input and previous output tokens. 
Even the only method applied to autoregressive S2T models \cite{kavaki20_interspeech} assumes conditional independence between prediction steps, disregarding 
the influence of previously generated text on each new prediction.
Finally, these methods are either incompatible with the spectrogram input format \citep{{markert-etal-2021,wu-etal-2023-explanations, wu-etal-2024-trust}}, which is typically used in modern S2T models,
or produce explanations that 
fail to highlight fine-grained patterns related to acoustic characteristics of speech, such as the
fundamental frequency and formants \citep{mandel16_bubble, trinh18_interspeech, kavaki20_interspeech, trinh21_bubble}.

To overcome the above limitations, we propose \textit{\textbf{S}pectrogram \textbf{P}erturbation for \textbf{E}xplainable \textbf{S}peech-to-text Generation} (\textbf{SPES}), the first feature attribution technique that provides token-level explanations for autoregressive S2T models by considering both the speech input and previously generated tokens.\footnote{The code is available at \url{https://github.com/hlt-mt/FBK-fairseq} under the Apache License 2.0.} 
SPES employs a perturbation-based approach that adapts image segmentation to spectrograms, enabling the identification of fine-grained, meaningful patterns.
Through quantitative assessments and in-depth analyses across two S2T tasks---ASR and, for the first time, speech translation (ST)---we show that SPES produces accurate, phonetically interpretable explanations aligned with human understanding.

\section{Background}
\label{sec:background}

Feature attribution methods produce a \textit{saliency} score vector $S = [s_1, s_2, \ldots, s_d]$, where each $s_i$ represents the contribution of input feature $x_i$ to the prediction $y$ \citep{samek-muller-2019, samek-etal-2021, madsen-etal-2022}. $S$ is typically visualized as a heatmap, or \textit{saliency map}, that highlights influential input regions. Existing methods fall into four categories: gradient-based, amortized, decomposition-based, and perturbation-based, each offering distinct ways to generate $S$.

\textit{Gradient-based} methods \citep{simonyan-etal-2014,sundararajan-etal-2017,Selvaraju-etal-2019} use the model's gradients with respect to input features to measure feature importance, where the gradient reflects the sensitivity of the prediction to small changes in each feature. While these methods are efficient, as they involve a single or a few backpropagation steps, they can be unstable due to gradient instabilities and sensitivity to noise \citep{ancona-etal-2018, samek-etal-2021, dombrowksi-etal-2022}.

\textit{Amortized explanation} methods \citep{dabkowski-gal-2017, yoon-etal-2019, schwarzenberg-etal-2021, chuang-etal-2023b} train a global explainer to produce saliency maps in a single forward pass. Although efficient at inference time, this approach requires separate training for each model and depends on the explainer’s learning capacity, which can be influenced by biased training patterns \cite{chuang-etal-2023}.

\textit{Decomposition-based} methods \citep{Bach-etal-2015, Shrikumar-etal-2017} propagate predictions backward through the network, layer by layer, decomposing the prediction score into contributions from individual neurons according to predefined rules. This process continues until the score is distributed across the input features. While these methods are as efficient as gradient-based ones, their effectiveness depends on the choice of propagation rules, which can vary across models and applications, potentially limiting their generalizability \citep{montavon-etal-2019}.

\textit{Perturbation-based} methods \citep{zeiler-fergus-2014, lundberg-etal-2017, fong-vedaldi-2017} determine feature saliency by systematically altering or masking parts of the input and observing the impact on the model's output. Although computationally intensive, especially for high-dimensional data, they provide stable and accurate explanations independent of model architecture \citep{covert-etal-2021}.

\section{Related Works}
\label{sec:related_works}

Compared to other fields, feature attribution in S2T tasks has received relatively little attention, with efforts focused only on ASR, leaving other tasks like ST unexplored. Moreover, some of the proposed methods do not base their explanations on spectrograms, which is the standard input representation for state-of-the-art S2T models. Specifically, \citet{markert-etal-2021} generate explanations based on mel-frequency cepstral coefficients, while \citet{wu-etal-2023-explanations, wu-etal-2024-trust} generate them directly on raw waveforms.

Henceforth we focus on those feature attribution techniques that work with spectrograms. Spectrograms provide a 2D visualization of frequency distributions over time, where darker areas indicate higher energy at specific frequencies, capturing the acoustic characteristics of speech \cite{stevens-acoustic}.
These techniques adopt either perturbation-based or amortized approaches.

The perturbation-based Bubble Noise technique introduced by \citet{mandel16_bubble} and \citet{trinh21_bubble} obscures
time-frequency patterns in the spectrogram using white noise, except in 
randomly positioned ellipsoidal bubble regions where the noise is attenuated, allowing speech characteristics to remain visible. 
To assess the impact of perturbations, this technique assigns a binary label to each predicted word based on whether the word changes under the perturbed input. Final saliency scores are calculated by correlating these labels with the noise mask, which tracks the amount of noise added to the spectrogram.
Bubble Noise involves high computational costs to generate explanations, which escalate with increasing bubble counts. Additionally, the use of uniformly sized and shaped bubbles inherently overlooks the structural patterns within the spectrogram, potentially limiting the method’s effectiveness in capturing fine-grained acoustic details.

To reduce explanation time and provide fine-grained insights, \citet{trinh18_interspeech} and \citet{kavaki20_interspeech} propose an amortized explanation method based on perturbations. A mask estimator is trained to regulate white noise addition to individual features in the spectrogram
aiming to maximize noise while preserving ASR accuracy. 
As a result, regions of the spectrogram where noise is minimized or absent are identified as salient for word prediction.
However, in addition to the high costs associated with training separate explainers for different models, the resulting explanations highlight time-frequency patterns that only loosely correspond to acoustically meaningful characteristics \citep{kavaki20_interspeech}.

Besides the above limitations, both techniques lack adaptation to the autoregressive nature of modern S2T models, where each token generation depends on both input speech and prior output text.
Leveraging the accuracy and architectural independence of perturbation-based methods, we fill these gaps by developing a technique that \textit{i)} accounts for the autoregressive nature of modern S2T models, and \textit{ii)} highlights fine-grained time-frequency patterns present in the spectrograms.

\section{SPES}
\label{sec:methodology}

Given an autoregressive S2T model that
generates one token at a time, conditioned on both the speech input $X$---represented as a spectrogram---and the previously generated tokens $Y^{(k-1)} = [y_0, y_1, \ldots, y_{k-1}]$,\footnote{Where $y_0$ is the special token start of sentence \texttt{<s>.}}
SPES produces two saliency maps: $\boldsymbol{S_{y_k}^{X}}$, which reveals the importance of input features across time and frequency, and $\boldsymbol{S_{y_k}^{Y^{(k-1)}}}$, which highlights the importance of previously generated tokens. The overall explanation for the predicted sequence $Y$, comprising $K$ tokens, is then formulated as:
\begin{equation}
S_Y = \{(S_{y_1}^{X}, S_{y_1}^{Y^{0}}), \ldots, (S_{y_K}^{X}, S_{y_K}^{Y^{(K-1)}})\}
\end{equation}

Preliminary experiments (detailed in Appendix \ref{app:token_validation}) indicated that separately perturbing the spectrogram and the previous output tokens yields more accurate saliency maps compared to joint perturbation. Therefore, we outline how these two components are perturbed independently to generate the corresponding saliency maps. 
Specifically, the following sections describe SPES in reference to
the key components of perturbation-based approaches \cite{covert-etal-2020b}, namely \textit{i)} selecting and perturbing features (\S \ref{sec:input_perturbation}), \textit{ii)} estimating the impact of these perturbations on model predictions (\S \ref{sec:spes_perturbation_impact}), and \textit{iii)} aggregating these impact estimations to derive final saliency scores (\S \ref{sec:aggregation}).

\subsection{Input Perturbation}
\label{sec:input_perturbation}

Input perturbation is usually performed by repeatedly masking input elements. 
However, perturbing each input element in combination with all others is impractical due to the exponential growth of possible combinations as the input size increases \cite{khakzar-etal-2020, covert-etal-2021}. Therefore, perturbation-based approaches typically sample a subset of combinations, often employing techniques like Monte Carlo sampling \cite{castro-etal-2017, petsiuk-etal-2018, markert-etal-2021, yang-etal-2021, fel-2023-perturbation}.
Accordingly, we sample spectrogram features and previous output tokens for perturbation $N^X$ and $N^Y$ times, respectively, with $N^X$ and $N^Y$ serving as hyperparameters that balance the accuracy of the saliency map against computational costs.

\subsubsection{Perturbing Spectrograms}
\label{sec:spectrogram_method}

\begin{figure}
    \centering
    \begin{subfigure}[b]{0.48\textwidth}
    \includegraphics[width=\textwidth]{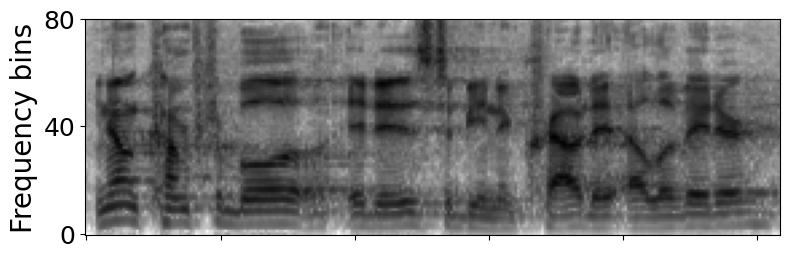}
    \end{subfigure}
    \begin{subfigure}[b]{0.48\textwidth}
    \includegraphics[width=\textwidth]{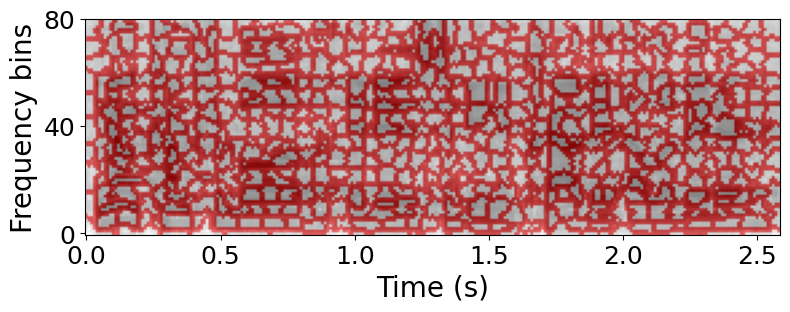}
    \end{subfigure}
    \caption{A mel-spectrogram (upper figure) and its segmentation obtained with SLIC (bottom figure).}
    \label{fig:segmentation}
\end{figure}

To achieve explanations that highlight fine-grained acoustic characteristics in spectrograms,
we design perturbations that target time-frequency patterns using techniques from image processing, where morphological clustering is often employed to define the perturbation units \citep{seo-etal-2020, yang-etal-2021}.\footnote{
In images, perturbations are typically applied either at the pixel level \citep{wagner-etal-2019} or to groups of pixels \citep{zeiler-fergus-2014, ribeiro-etal-2016, yang-etal-2021}, referred to as \textit{patches} or \textit{superpixels}, which serve as the minimal processing units.}
This approach produces explanations that align more naturally with object contours \citep{ivanovs-etal-2021}, in contrast to perturbations with uniformly sized and shaped units.
A particularly effective clustering-based technique is 
the \textit{Morphological Fragmental Perturbation Pyramid} (MFPP; \citealt{yang-etal-2021}), which generates detailed saliency maps \citep{ivanovs-etal-2021}. MFPP employs Simple Linear Iterative Clustering (SLIC; \citealt{achanta-etal-2012}), a \textit{k}-means-based algorithm that groups pixels according to spatial and color characteristics---representing, in our case, proxies for spectral patterns.
By iteratively adjusting the number of clusters (\textit{k}), the method produces a hierarchy of patches across varying granularities, from fine to coarse.

As shown in Fig. \ref{fig:segmentation}, our SLIC-based segmentation of the spectrogram clusters nearby features with similar energy levels into distinct patches that respect the structure of acoustic patterns, such as formants.
Following MFPP, each spectrogram is segmented at three levels of granularity by varying \textit{k}.
This \textit{multi-scale} approach creates patches at different resolutions, enabling the perturbation of neighboring features together or separately.

However, unlike images that have fixed dimensions, speech data vary in duration.  Consequently, a fixed value of \textit{k} would result in larger patches for longer audio samples, leading to coarser explanations. To address this, SPES scales \textit{k} proportionally to input length, ensuring consistent patch sizes across varying speech sample lengths. 
However, this approach has a limitation: as the number of patch combinations grows exponentially with \textit{k}, exhaustive exploration of the search space requires a corresponding increase in the number of perturbations $N^X$, which becomes computationally intractable for large values of \textit{k}.\footnote{As shown in Fig. 
\ref{fig:time} in Appendix \ref{app:complementary-results}, the relationship between audio duration and computation time for generating explanations is quadratic, meaning that longer audio samples account for most of the computation time. Hence, increasing $N^X$ with the audio length would substantially increase computational costs.}
Insufficient exploration, on the other hand, eventually lowers explanation quality (see Appendix \ref{app:tau}). To balance computational cost and effective exploration of patch combinations, SPES limits the increase of \textit{k} by setting a maximum value. Specifically, with $len(X)$ as the input length in seconds, $\phi$ as the number of patches per second, and $\tau$ as the threshold length in seconds, \textit{k} is determined as:

\begin{equation}
k = \text{min}(len(X), \tau) \cdot \phi
\label{eq:duration_adaptation}
\end{equation}

Once the patches are defined, we randomly perturb them in combination across $N^X$ iterations. 
The number of patches perturbed at each iteration is controlled by a hyperparameter ($p_{spec}$), which specifies the probability of perturbing each patch---for example,
with $p_{spec}$ = $0.5$, approximately half of the patches are perturbed.

Perturbation is implemented by setting the elements of the patch to $0$
in a binary mask $M^X \in \{0,1\}^{T \times C}$, where $T$ represents
the time dimension and $C$ represents
the frequency dimension of $X$.
The perturbed input $X'$ is then obtained by element-wise multiplication of the original input $X$ and $M^X$, which corresponds to setting the perturbed features to $0$, 
the mean energy value in the utterance.\footnote{In line with standard practices in current S2T systems \citep[][among others]{seamless}, all input spectrograms undergo cepstral mean-variance normalization  \citep{VIIKKI1998133} before being fed into the model.}
This perturbation method---the same used at training time by SpecAugment \citep{specaugment}, a common augmentation technique in most training settings, including ours---avoids introducing artifacts that fall outside the training distribution, helping the model remain robust to perturbations and aligning with strategies to prevent hallucinations during perturbations \citep{Holzinger2022, brocki-chung-2023}.

\subsubsection{Perturbing Previous Output Tokens} 
\label{subsec:token_method}

Inspired by the effectiveness demonstrated in 
context mixing by
the \textit{value zeroing} \citep{mohebbi-etal-2023} perturbation strategy, which replaces selected token embeddings with zero vectors, SPES adopts the same approach for the perturbation of the previous output tokens. Specifically, for each of the $N^Y$ iterations, it builds a binary mask $M^{Y^{(k-1)}} \in \{0, 1\}^{(k-1) \times 1}$, where each token is randomly selected for perturbation with probability $p_{tok}$, and the corresponding mask values are set to 0. The mask $M^{Y^{(k-1)}}$ is then multiplied with the embedding matrix of the previous output tokens to obtain their perturbed version $Y^{(k-1)'}$.

\subsection{Perturbation Impact Estimation}
\label{sec:spes_perturbation_impact}

The importance of the features perturbed at each iteration is assessed by estimating how the perturbation impacts the model’s ability to predict the output. 
In classification tasks, two common methods are used for this estimation, both based on the probability assigned by the model to the target class. One method uses the target class probability as the saliency score for non-perturbed features \citep{zeiler-fergus-2014, petsiuk-etal-2018, yang-etal-2021},
where a high score indicates that the perturbed features do not include information relevant to predicting the target class.
Alternatively, the score is computed as the difference between the original probability (without perturbation) and the probability obtained with the perturbed input \citep{ancona-etal-2018, seo-etal-2020}. In this case, the score is assigned to the perturbed features, 
where a high score suggests that the model’s ability to recognize the target class is compromised in their absence.

In S2T tasks, assessing single-class probabilities corresponds to using the probability of the token being explained. 
We posit that this approach can be improved for two reasons.
First, while classification tasks typically deal with a limited number of target classes, S2T tasks involve a BPE vocabulary \citep{sennrich-etal-2016-neural} with thousands of tokens \citep{di-gangi-etal-2020-target,fendji-etal-2022}. 
This results in output probability distributions that are high-dimensional, dense vectors rather than sparse, peaked distributions \citep{zhao-etal-2023}. 
Second, the decoding process of S2T models often employs beam search \citep{NIPS2014_a14ac55a}, where the tokens with the highest probability are not always selected, and perturbations can significantly affect the predicted sequence by altering the probabilities of initially-neglected tokens.

Therefore, drawing on mechanistic interpretability research \citep{conmy-etal-2023, zhang-nanda-2024}, SPES estimates perturbation impact ($r^X$ for spectrograms and $r^{Y^{(k-1)}}$ for previous output tokens) by computing the KL divergence \citep{10.1214/aoms/1177729694} between 
the probability distribution of an S2T model $\Psi$ fed with the original inputs ($X$ and $Y^{(k-1)}$) and that obtained with their perturbed counterparts ($X'$ and $Y^{(k-1)'}$).
Specifically:
\begin{equation*}
\begin{aligned}
   r^X &= \text{KL}\left(\Psi(X, Y^{(k-1)}), \Psi(X', Y^{(k-1)})\right) \\
   r^{Y^{(k-1)}} &= \text{KL}\left(\Psi(X, Y^{(k-1)}), \Psi(X, Y^{(k-1)'})\right)
\end{aligned}
\end{equation*}

\subsection{Scores Aggregation}
\label{sec:aggregation}

Following MFPP, the $r$ scores are aggregated
across all perturbation iterations to produce the saliency maps:
\begin{equation}
S = \frac{\sum_{i=1}^{N} r_i (1-M_i)}{\sum_{i=1}^{N} (1-M_i)}
\label{eq:saliency_map}
\end{equation}
where $S$ is either $S^X$ or $S^{Y^{(k-1)}}$ and, accordingly, $M$ is either $M^X$ or $M^{Y^{(k-1)}}$, $N$ is either $N^X$ or $N^Y$, and $r$ is either $r^X$ or $r^{Y^{(k-1)}}$.
Specifically, Eq. \ref{eq:saliency_map} sums the $r$ scores assigned to the perturbed features ($1 - M_i$) in the numerator and weights them by the number of times each feature is perturbed in the denominator.

Lastly, SPES applies two normalization steps. 
First, it applies z-score normalization to each token-level saliency map ($S_{y_k}^{X}$ and $S_{y_k}^{Y^{(k-1)}}$) independently.
This step addresses the varying magnitudes of saliency scores across different tokens $y_k$ in the same output $Y$, 
preventing disproportionate emphasis on specific tokens.
This is especially beneficial when aggregating token-level explanations at the sentence level, such as through averaging, to create a single saliency map that reflects the importance of features for the entire predicted sentence.
The effectiveness of this normalization has been validated on the validation set, as detailed in Appendix \ref{app:spectrogram_validation}. Second, SPES applies joint min-max normalization to $S_{y_k}^{X}$ and $S_{y_k}^{Y^{(k-1)}}$ for the same token $y_k$, scaling their scores to the $[0, 1]$ range. This paired normalization ensures interaction between the two maps: if one map has higher scores for a token, the other map will correspondingly receive lower scores, thereby reflecting the relative importance of $S_{y_k}^{X}$ and $S_{y_k}^{Y^{(k-1)}}$ for predicting $y_k$.

\section{Evaluating Explanations in S2T}
\label{sec:evaluation}

Evaluating the quality of explanations is a complex task, ideally involving the assessment of several desirable properties across three dimensions: content, presentation, and user \citep{nauta-etal-2023}.
Within the \textbf{content} dimension, the foremost property is \textit{correctness}, also known as \textit{fidelity} \citep{tomsett-etal-2020} or \textit{faithfulness} \citep{jacovi-goldberg-2020}, which refers to ``how well explanations match model reasoning'' \citep{li-etal-2023}.
Moving to the \textbf{presentation} dimension, the quality of an explanation concerns not only its composition (``how something is explained'') but also its \textit{compactness}, adhering to the ``less is more'' principle.
Accordingly, explanations should highlight the smallest possible set of features and avoid the inclusion of redundant elements \citep{sun-etal-2020}.
Lastly, concerning the \textbf{user} dimension, a desirable property is \textit{plausibility}, defined as ``how convincing the explanation is to humans'' \citep{jacovi-goldberg-2020}.

In this work, we consider all three dimensions relevant to evaluating explanations. We inspect the content dimension by assessing faithfulness with the \textit{deletion} metric, a popular method in classification tasks \citep{arras-etal-2017, samek-etal-2017, petsiuk-etal-2018, tomsett-etal-2020, samek-etal-2021, Gevaert2024} and also used in ASR \cite{trinh20_interspeech}. On the presentation axis, we measure compactness using the \textit{size} metric. Finally, on the user axis, we assess plausibility through \textit{dedicated analyses} discussed in \S\ref{sec:plausibility}.

\paragraph{Deletion}
This metric evaluates how quickly performance declines as input features, ranked by saliency scores, are gradually removed.
This is shown through a curve, where a steeper drop indicates that the most important features were removed early. The area under the curve (AUC) reflects how well the explanation identifies key features---the lower the AUC, the better.
\citet{trinh20_interspeech} applied deletion to ASR explanations, adding noise to the most relevant features for each word in the output sentence. 
Thereby, they test whether the prediction of the target word changes under the noisy condition, while predictions for other words remain unaffected, since the noisy features are not expected to influence them. The process is repeated across various subsets of features with different saliency levels.
However, this approach assumes conditional independence in predicting different words, an assumption that does not hold in state-of-the-art autoregressive S2T models, where early errors can impact subsequent predictions.
To address this limitation, we adapt the deletion metric to remove features from the spectrogram based on their saliency at the sentence level and evaluate performance on the entire sentence using standard ASR and ST metrics as measures of accuracy.
Specifically, we incrementally zero out the most salient features at the sentence level from the spectrogram in steps of $5\%$
and assess downstream performance across the entire test set. We use WER\footnote{Computed with editdistance 0.6.1 (\url{https://github.com/roy-ht/editdistance}). Scores are capped at $100$ to mitigate instability from excessively high scores due to long hallucinations.} for ASR and BLEU\footnote{Computed with sacreBLEU 1.5.1 \citep{post-2018-call}.}
\citep{papineni-etal-2002-bleu} for ST, and then measure the deletion AUC. Since WER decreases with better performance, while BLEU increases, faithful explanations should yield high AUC values for ASR and low AUC values for ST.

\paragraph{Size}
Size measures the compactness of an explanation as the percentage of input features identified as responsible for the model's prediction. In classification tasks, this involves incrementally adding the features with the highest saliency scores until the model's prediction matches the original \citep{sun-etal-2020}. However, applying this approach in S2T tasks is complex since the output text changes continuously rather than in discrete steps as in classification. 
\citet{wu-etal-2024-trust} test two criteria to detect changes: \textit{i)} WER greater than $0$, which considers any modification---even irrelevant---as a prediction change, and \textit{ii)} cosine similarity of SentenceBERT encodings lower than $0.5$, which targets changes in output semantics. Since  the choice of the
similarity metric can lead to varying results and the optimal metric is debatable, we avoid isolating the subset of most salient features; instead, we assess explanation size across various subsets.
To this aim, we calculate the percentage of features with saliency scores above a threshold ranging from $0$ to $1$ in $5\%$ increments and compute the AUC of the resulting curve as a comprehensive score. 
Lower AUCs reflect more compact explanations, resulting in more concise and clear presentation \citep{nauta-etal-2023}.

\section{Experimental Setting}
\label{sec:experimental_setting}

\paragraph{Tasks and Data} 
We test SPES on S2T tasks with English audio, focusing on ASR and ST into German (en$\rightarrow$de) and Spanish (en$\rightarrow$es) to assess its effectiveness across varying monotonicity in word alignment.\footnote{English and Spanish follow the SVO order, while German is an SOV language \citep{ostling-2015}.} 
For evaluation, we use the MuST-C corpus \citep{mustc}: the en$\rightarrow$de v2 subset for ASR and en$\rightarrow$de ST, and the en$\rightarrow$es v1 subset for en$\rightarrow$es ST due to the absence of v2.

\paragraph{Models} 
The ASR and two ST systems share the same architecture, consisting of a Conformer encoder and a Transformer decoder, and are fed with 80-feature mel-spectrograms (see Appendix \ref{appendix:exp_setting} for details on hyperparameters, training configurations, and data). We opted not to use widely adopted foundation models like Whisper \citep{whisper} or SeamlessM4T \citep{seamless} for several reasons: \textit{i)} comparability across tasks, as Whisper only supports English translations; \textit{ii)} performance, as our models outperform SeamlessM4T on these tasks (see comparison in Appendix \ref{appendix:exp_setting}); and \textit{iii)} computational efficiency, since our smaller models are more suited for extensive experiments while keeping computational costs manageable.

\paragraph{SPES Hyperparameters} 
SPES hyperparameters were estimated through validation (see Appendix \ref{app:spes_validation}). Spectrograms were clustered into patches using SLIC with default parameters from the scikit-image implementation,\footnote{\url{https://scikit-image.org/docs/dev/api/skimage.segmentation.html}} except for sigma ($\sigma$), which was set to $0$. We set $\phi$ to \num{400}, \num{500}, and \num{600} for the three-level multi-scale clustering, while $\tau$ was set to $7.5$ for ASR and $5$ for ST. The number of masking iterations, $N^X$, was set to \num{20000} with $p_{spec}$ at $0.5$. For previous output tokens, $N^Y$ was set to \num{2000} with $p_{tok}$  at $0.4$.

\paragraph{Baselines}
We compare SPES against two baselines. First, a \textit{\textbf{feature-wise}} method that randomly perturbs previous output tokens and spectrograms at the level of single features without morphological clustering, and estimates the perturbation impact with KL divergence.  For a fair comparison, we set $N^X$ to \num{20000} and $N^Y$ to \num{2000}, as in SPES, with $p_{spec}$ = $0.7$ and $p_{tok}$ = $0.1$, optimized via validation (see Appendix \ref{app:features_wise_validation}). Second, an enhanced version of the \textit{\textbf{Bubble Noise}} technique that uses KL divergence as a perturbation impact estimator and is extended to also perturb the previous output tokens. Since Bubble Noise is significantly slower than  SPES, especially at high bubble counts, we set $N^X$ to \num{1000}.  This value  corresponds to that used by \citet{trinh20_interspeech} and results in a runtime comparable to SPES with $N^X$ set to \num{20000}. The number of bubbles per second is set to $10$, with each bubble measuring approximately $0.43$ seconds in width and $31$ mel bins in height, as validated in Appendix \ref{app:bubble}.

\begin{table*}[!t]
\small
\centering
\begin{tabular}{l|cc|cc|cc}
\toprule
 & \multicolumn{2}{c|}{\textbf{ASR (en)}} & \multicolumn{2}{c|}{\textbf{ST (en$\rightarrow$de)}} & \multicolumn{2}{c}{\textbf{ST (en$\rightarrow$es)}} \\
\cline{2-7}
 & \textbf{Deletion} ($\uparrow$) & \textbf{Size} ($\downarrow$) & \textbf{Deletion} ($\downarrow$) & \textbf{Size} ($\downarrow$) & \textbf{Deletion} ($\downarrow$) & \textbf{Size} ($\downarrow$) \\
\hline
feature-wise & 71.39 & 47.85 & 8.79 & 48.53 & 9.03 & 46.00 \\
Bubble Noise & 65.02 & 36.92 & 10.85 & 43.89 & 13.01 & 43.04 \\
\midrule
SPES & \textbf{92.54} & \textbf{29.71} & \textbf{2.34} & \textbf{29.28} & \textbf{2.45} & \textbf{29.47} \\
\hspace{4mm}– multi-scale & 92.22  & 29.94 & 2.49 & 29.62 & 2.56 & 29.78 \\
\hspace{4mm}– clustering & 90.73 & 33.31 & 2.84 & 33.09 & 2.93 & 32.90 \\
\hspace{4mm}– duration adaptation & 92.37 & 30.45 & 2.35 & 31.10 & 2.46 & 30.95 \\
\hspace{4mm}– KL & 87.44 & 29.85 & 4.76 & 30.98 & 4.99 & 30.74 \\
\bottomrule
\end{tabular}
\caption{Deletion and Size scores for the \textit{feature-wise} and the \textit{Bubble Noise} baselines as well as for SPES and the solutions tested in the ablation study.}
\label{tab:results}
\end{table*}

\section{Results}
\label{sec:results}

\subsection{Overall Results}

The upper rows of Table \ref{tab:results}  present 
Deletion and Size AUC scores  
for the two implemented baselines (\textit{feature-wise} and \textit{Bubble Noise}) and compare them with SPES across ASR (en) and ST (en$\rightarrow$de, en$\rightarrow$es).
Full deletion and size curves are provided in Fig. \ref{fig:deletion} and \ref{fig:size} in Appendix \ref{app:complementary-results}.

Deletion and size scores reveal consistent trends across the three tasks. Looking at the baselines, we observe that the \textit{feature-wise} method outperforms \textit{Bubble Noise} in deletion scores, likely due to the differing number of perturbation iterations ($N^X$) assigned to the two baselines (see \S \ref{sec:experimental_setting}). 
However, despite the significantly lower $N^X$, \textit{Bubble Noise} surpasses \textit{feature-wise}
in size scores.

When comparing SPES to the baselines, it clearly excels. In terms of deletion scores, SPES outperforms the \textit{feature-wise} baseline by $21.15$ points in ASR and up to $6.58$ points in ST (en$\rightarrow$es). The improvements are even more pronounced relative to the \textit{Bubble Noise} technique ($27.52$ points in ASR and $10.56$ in ST en$\rightarrow$es). SPES also demonstrates its superiority in
the size metric, with differences of up to $19.25$ and $14.61$ in ST (en$\rightarrow$de) compared to \textit{feature-wise} and \textit{Bubble Noise}, respectively.
Since the comparison between SPES and \textit{Bubble Noise} involves different $N^X$ values, for the sake of fairness we carried out additional experiments using the same $N^X$ value (\num{1000}) for both SPES and \textit{Bubble Noise}. The results, presented in Table \ref{tab:results_bubble} in Appendix \ref{app:complementary-results}, confirm SPES's superior performance under this condition as well.
\finding \hspace{0.1 mm} Overall, \textbf{SPES demonstrates its capability to generate faithful and compact explanations}, accurately highlighting the input elements that are relevant to the S2T model's predictions while avoiding the inclusion of redundant information.

\subsection{Ablation Study}
\label{sec:ablation}

To better understand how each design choice impacts the faithfulness and compactness of SPES-generated explanations, we conduct an ablation study. Firstly, we vary the granularity of spectrogram perturbations by testing: \textit{i)} single-level segmentation instead of multi-level (\textit{– multi-scale}), by using only \num{500} as $\phi$, 
\textit{ii)} fixed grid segmentation\footnote{Obtained setting the sigma parameter in SLIC, which determines the smoothness of the segmentation contours, to $10$.} instead of clustering (\textit{– clustering}), \textit{iii)} removal of the duration-based patch adaptation of Eq. \ref{eq:duration_adaptation} (\textit{– duration adaptation}), using a fixed number of patches (\num{2000}, \num{2500}, and \num{3000})\footnote{Validated on the dev set, as described in Appendix \ref{app:spectrogram_validation}.} for all audio samples regardless of their duration. Lastly, we assess the contribution of KL divergence by replacing it with token-level probability difference---widely used in text applications \citep{sarti-etal-2023-inseq, miglani-etal-2023-using}---as a method for estimating perturbation impact (\textit{– KL}). 

Table \ref{tab:results} (lower rows) presents the scores resulting from these variations. Looking at deletion, the largest performance drop across all tasks occurs when KL divergence is replaced with token-level probability difference (\textit{– KL}). Ablating KL divergence reduces performance also in terms of size, although differences are less evident (ranging from $0.14$ to $1.70$ points).
\finding~\hspace{0.1 mm}~We can conclude that \textbf{using KL divergence to estimate perturbation impact is beneficial for producing explanations that are both more faithful and more compact}.
Examining the ablations related to patch definition in the input spectrogram, we find that considering morphological aspects has the greatest impact.
Indeed, fixed grid segmentation (\textit{– clustering}) leads to decreased faithfulness for both ASR and ST, along with a $12$-$13\%$ increase in explanation size across all tasks. 
\finding~\hspace{0.1 mm}~This indicates that \textbf{accounting for morphology in spectrograms is crucial} for isolating patterns that models exploit for their predictions.
\finding~\hspace{0.1 mm}~\textbf{Adapting the number of patches to the input speech duration also proves beneficial}: 
with a fixed number of patches (\textit{– duration adaptation}), deletion scores are slightly worse but similar, and the size increases by $2$-$6\%$. 
\finding~\hspace{0.1 mm}~Lastly,m \textbf{using multiple segmentation levels improves both faithfulness and compactness}: ablating this option (\textit{– multi-scale}) has a slight negative impact across all tasks, reflected in both deletion and size scores, with a degradation of approximately $1\%$ in size for ST.

In summary, all the design choices of SPES contribute to providing faithful and compact explanations. Among them, estimating the impact of perturbations with KL divergence proves to be particularly crucial for achieving meaningful outcomes, followed by employing morphological clustering to define the units for perturbation.

\section{Analyses of Plausibility}
\label{sec:plausibility}

We assess the \textit{plausibility} of SPES explanations by analyzing the saliency maps for both spectrograms ($\boldsymbol{S^X}$) and previously generated tokens ($\boldsymbol{S^{Y^{(k-1)}}}$). For $S^X$, we focus on ASR, where the more monotonic relationship between audio input and output tokens facilitates the analysis. For $S^{Y^{(k-1)}}$, we examine both ASR and ST.

\subsection{Explanations for Spectrograms}
\label{sec:plausibility_spectrogram}

\paragraph{Time Dimension} 
Especially in ASR, manual inspection of $S^X$ examples suggests that features' saliency moves monotonically along the time dimension as the generation progresses from the first to the last token (for an example, see Appendix \ref{app:examples}). This observation aligns with the expectation that the generation is guided by the time frames in which the token is uttered, so explanations should be time-localized and shift from the start to the end of the audio sample. 
We verify this quantitatively by comparing word-level saliency maps on the time axis\footnote{To this aim, we first aggregate explanations at the word level by averaging the saliency maps of the tokens belonging to the same word, and then we take the maximum value over the frequency axis for each time frame.} with word-level alignments provided by Gentle.\footnote{\url{https://github.com/lowerquality/gentle}.}
Specifically, we compare the mean score of the saliency maps for time frames \textit{inside} and \textit{outside} the boundaries indicated by Gentle, and test with a Student's T-test the hypothesis that the mean score inside the time span is higher than that outside of the time span.  
\finding \hspace{0.1 mm} 
As the test rejects the null hypothesis that the two means are equal with $p<0.01$, we can assert that 
\textbf{SPES offers plausible insights on the time dimension.}

\paragraph{Frequency Dimension}
To assess whether SPES can also offer plausible insights along the frequency dimension, we analyze the distribution of explanation scores related to: \textit{i)} phones exhibiting different spectral characteristics, and \textit{ii)} same phones uttered by men and women.

In the first case, we select a \textit{minimal pair} \citep{bale-reiss-2023} of words that differ in a single phoneme: \textit{so} (/\textipa{so\textupsilon}/) and \textit{no} (/\textipa{no\textupsilon}/).\footnote{We refer to the standard American English pronunciation, but we acknowledge that the actual pronunciation in individual samples may vary due to sociophonetic factors, including dialectal variations among speakers.} 
The former features a voiceless alveolar sibilant /s/, characterized by turbulent airflow with high-frequency energy, typically above \num{4000} Hz. The latter contains a voiced alveolar nasal /n/, exhibiting a low-frequency murmur, typically around $250$-$300$ Hz \citep{keith-2011}.
To check whether SPES highlights these specific characteristics, we select the explanations corresponding to the frames where these words are uttered, collect their maximum scores on the frequency axis, and  average these values.\footnote{The number of occurrences for \textit{so} is $143$, for \textit{no} is $46$.}
Results are plotted in Fig. \ref{fig:freq} (top) in Appendix \ref{app:complementary-results}. Several peaks over $0.5$ points are noticeable. Some of them (e.g., around $500$ and \num{1200} Hz) are common  to the two words, and reflect the typical frequency characteristics of back rounded vowels like those in the diphthong /\textipa{o\textupsilon}/ \citep{hillenbrand-etal-1995}. 
Others, instead, are peculiar of the two distinct phones: for \textit{so}, scores are high above \num{4000} Hz, as expected for alveolar sibilants, whereas \textit{no} shows a prominent peak in the lower frequency region, around $300$ Hz, corresponding to the murmur of an alveolar nasal (for examples of these saliency maps, see Fig. \ref{fig:so} and \ref{fig:no} in Appendix \ref{app:complementary-results}).

In Fig. \ref{fig:freq} (bottom) in Appendix \ref{app:complementary-results}, we compare saliency scores for men's and women's utterances of \textit{so}. Due to the typically higher frequencies in women's voices, linked to shorter vocal tract and smaller larynx size \citep{stevens-acoustic}, SPES's saliency scores highlight slightly higher frequencies for women's utterances compared to men's.

\finding \hspace{0.1 mm} These analyses indicate that \textbf{SPES can offer valuable insights into the frequency dimension}, capturing phonetic differences between sounds and sociophonetic characteristics such as gender.

\subsection{Spectrograms vs Previous Output Tokens}
\label{subsec:spectogram-output}
Previous analyses show that SPES-generated saliency maps generally localize salient areas effectively in both time and frequency domains. However, manual inspections revealed that some spectrogram saliency maps appear ``noisy'', with scattered explanations that lack concentrated groups of features with higher saliency scores (for an example, see Appendix \ref{app:examples}).
To investigate this behavior further, we explore whether this scattering is associated with specific token classes. Specifically, we hypothesize that, for some tokens, the  less compact
explanations in terms of speech features may stem from a higher contribution of previous output tokens to their prediction.

To quantify scattering, we use kurtosis, a measure of the ``tailedness'' of score distributions. Higher kurtosis indicates more peaked distributions, reflecting more focused explanations.
For each token occurring over $100$ times in the test set, we average the kurtosis values of all the spectrogram saliency maps related to its occurrences.

Interestingly, a trend emerges: tokens with higher kurtosis generally correspond to full words,  while those with lower kurtosis are punctuation marks or special tokens. Fig. \ref{fig:kurtosis} shows the top and bottom $5$ tokens by kurtosis.  The bottom tokens include punctuation marks, the special end-of-sentence token \texttt{</s>}, and  \texttt{\_years}. The scattered explanations for these tokens suggests a lack of crucial audio information for their prediction. This is expected for punctuation, which often depends on grammatical rules rather than audio content. Similarly, \texttt{</s>} is always predicted based on a preceding strong punctuation mark, independent of the audio. In the case of \texttt{\_years}, the model often relies on a preceding number in the output tokens, indicating a pattern-based prediction.

\finding \hspace{0.1 mm} These findings reveal that \textbf{SPES can be used to determine whether a token is predicted based on the input speech or on the textual context}, as it highlights the dependency of a predicted token on previous ones only when plausible. In the next section, we further explore the plausibility of the textual relationships unveiled by SPES.

\subsection{Explanations for Previous Output Tokens}
\label{subsec:analyses_tokens}

\paragraph{Positional Saliency}
A manual inspection of $S^{Y^{(k-1)}}$ in both ASR and ST revealed that two types of tokens typically receive higher saliency scores: the special start-of-sentence token \texttt{<s>} and the latest token (\texttt{LT}), i.e., the token immediately preceding the one being explained (for an example, see Appendix \ref{app:examples}). Plotted distributions of the saliency scores for these token types and any  tokens in intermediate positions (\texttt{IT}) reflect our qualitative insights: \texttt{LT}s have the highest mean (see Fig. \ref{fig:token_position} in Appendix \ref{app:complementary-results}), followed by  \texttt{<s>}, whose distribution shifts towards higher scores compared to \texttt{IT}s. Student's T-tests, comparing the means of these groups, always reject the null hypothesis of equal means with $p<0.01$.

The importance of \texttt{LT}s underscores the critical role of the closest linguistic context. This finding is line with prior work that, by analyzing model component importance, demonstrated the significant focus of attention heads on \texttt{LT}s \citep{ferrando-etal-2024-flow} and of final token states in terminal layers for predicting the subsequent tokens \citep{meng-etal-2024}. Additionally, the significance of \texttt{<s>} echoes findings from research on ``attention sinks'' in autoregressive language models, which suggest that initial tokens act as anchors due to their positional values \citep{xiao-etal-2023}.

\finding \hspace{0.1 mm} In summary, concerning $S^{Y^{(k-1)}}$,  \textbf{SPES highlights mechanisms that align with findings from previous research}.

\paragraph{Intermediate Tokens}
Despite the aforementioned trends in
positional saliency, we observed instances where an \texttt{IT} holds the highest saliency score for the explanations of certain tokens (see  Appendix \ref{app:examples}). To investigate this phenomenon, we analyze tokens occurring over $100$ times in the test data for
each task. For each unique token, we calculate the frequency with which an \texttt{IT} received the highest score in the explanation.

Table \ref{tab:intermediate_token} presents examples from each language and task, showing the tokens where an \texttt{IT} held the highest score in at least $25$\% of their occurrences, along with their top two most frequent \texttt{IT}s, if available. Overall, these explanations appear plausible and linguistically motivated. First, punctuation marks are present across all languages. The token  \texttt{)} is explained by its complementary \texttt{(}. Similarly, \texttt{?} is explained by \texttt{What}/\texttt{what} in English, and by \texttt{¿} in Spanish, i.e. by \texttt{IT}s that typically introduce questions. This signals that the preceding context plays a pivotal role in explaining token pairs that frequently co-occur. Also, it aligns with our findings on punctuation marks (\S \ref{subsec:spectogram-output}): being characteristic of written text, punctuation does not have a direct correspondence in speech, reducing the saliency of the audio signal in the explanation.

By specifically looking at en$\rightarrow$de, the token \texttt{zu} is preceded by the \texttt{IT} \texttt{um} to create infinitive sentences. This German grammatical construction corresponds to the single token ``to'' in English (e.g., ``\textit{\textbf{\texttt{um}} Menschen aus ihren Stühlen \textbf{\texttt{zu}} holen}'' -- ``\textbf{to} get people out of their chairs.''). Along the same line, the token \texttt{an} acts as a separable prefix for the verb \texttt{fängt} (e.g. ``\textit{Es \textbf{\texttt{fängt}} später \textbf{\texttt{an}}}'' -- ``\textit{it \textbf{starts} later}''). Also, \texttt{an} is part of fixed expressions like ``\textit{es fühlt \textbf{\texttt{sich}} so \textbf{\texttt{an}}}'' (``\textit{it feels \textbf{like}}''). 
\finding \hspace{0.1 mm} These examples further demonstrate how \textbf{SPES relies on previous context to accurately interpret and explain tokens, particularly when there is no one-to-one mapping between source sentence and target translation.}

\section{Conclusions}

In response to the need for feature attribution methods specifically designed for S2T autoregressive models, this work introduces  \textbf{SPES} (\textit{\textbf{S}pectrogram \textbf{P}erturbation for \textbf{E}xplainable \textbf{S}peech-to-text Generation}). By employing a perturbation-based approach that ensures reliable explanations while remaining compatible with any model architecture, SPES produces token-level explanations throughout the predicted sentence. Its saliency maps are the first to effectively highlight contributions from fine-grained, acoustically meaningful patterns, such as formants, while also accounting for the influence of previous output tokens. We evaluate SPES on ASR and---for the first time---ST, and support this evaluation with in-depth analyses. Results show that SPES provides 
explanations that are faithful, compact, and plausible. This work contributes to enhancing the explainability of S2T models, laying groundwork for more transparent and reliable speech technology.

\bibliography{main}
\bibliographystyle{main}

\clearpage

\appendix

\section{Experimental Setting}

\subsection{Model Implementation}
\label{appendix:exp_setting}

In this section, we provide further details on the implementation of the ASR and ST models (see \S \ref{sec:experimental_setting}) used in our experiments.

\paragraph{Models}
Our models were implemented using the fairseq-S2T framework \citep{wang-etal-2020-fairseq}. As audio features, we use log-compressed mel-filterbanks ($80$ channels), computed over windows of $25$ms with a stride of $10$ms using pykaldi.\footnote{\url{https://github.com/pykaldi/pykaldi}.}
We encode text into BPE \citep{sennrich-etal-2016-neural} using SentencePiece \citep{kudo-richardson-2018-sentencepiece} with a vocabulary size of \num{8000} for English and \num{16000} for Spanish and German. The length of audio features is reduced by a factor of $4$ using two $1$D convolutional layers with stride $2$. The output of the convolutions is fed to a $12$-layer Conformer~\citep{gulati20_interspeech} encoder, and a $6$-layer Transformer~\citep{vaswani} decoder. We used $512$ embedding features, \num{2048} hidden features in the FFN, and a kernel size of $31$ for Conformer convolutions. In total, the models have $113$M parameters. As an objective, we minimize label-smoothed cross entropy \citep{szegedy-etal-2016} on the output of the decoder (using as a reference either the textual translation, for ST, or the transcripts, for ASR) with an auxiliary CTC \citep{Graves2006ConnectionistTC} loss with the transcripts as reference, regardless of the target task, computed on the output of the 8th encoder layer. We optimize the model weights with the Adam optimizer \citep{kingma-ba-2015} ($\beta_1=0.9$, $\beta_2=0.98$) and Noam learning rate scheduler \citep{vaswani} (inverse square-root) starting from $0$ and reaching the $0.002$ peak in \num{25000} warm-up steps. The input features are normalized with Utterance-level Cepstral Mean and Variance Normalization and, at training time, we apply SpecAugment \citep{specaugment}. We set dropout probability to $0.1$. The ASR model was trained for \num{250000} steps while the ST models, one for each language direction, were trained for \num{200000} steps, as their encoder was initialized with that of the ASR model. All trainings were carried out on four NVIDIA A100 GPUs ($40$GB of RAM) with \num{40000} tokens per mini-batch and $2$ as update frequency. All models were obtained by averaging the last 7 checkpoints.

\paragraph{Data}
Our ASR model is trained on the following datasets: CommonVoice \citep{commonvoice}, LibriSpeech \citep{librispeech}, TEDLIUM v3 \citep{tedlium}, and VoxPopuli \citep{voxpopuli}. 
For ST, instead, as training corpora we used MuST-C \citep{mustc}, EuroParl-ST \citep{europarl}, and CoVoST v2 \citep{covost}.
Additionally, we translated the transcripts of the ASR datasets into the two target languages with the NeMo MT models.\footnote{\url{https://docs.nvidia.com/nemo-framework/user-guide/latest/nemotoolkit/nlp/machine_translation/machine_translation.html}.}

Table \ref{tab:asr_st_res} reports the results of our models and compares them with SeamlessM4T \citep{seamless}.

\begin{table}[!ht]
\centering
\small
\begin{tabular}{l||c|cc}
\toprule
& \textbf{ASR}
 & \textbf{ST en$\rightarrow$de} & \textbf{ST en$\rightarrow$es} \\
 & \textbf{WER ($\downarrow$)}
 & \textbf{BLEU ($\uparrow$)} & \textbf{BLEU ($\uparrow$)} \\
\hline
\hline
Ours & \textbf{10.8} & \textbf{30.7} & \textbf{34.9 }\\
Seamless & 15.8 & 23.0 & 31.8 \\
\bottomrule
\end{tabular}
\caption{ASR (WER) and ST (BLEU) results of our models compared to SeamlessM4T on the MuST-C tst-COMMON set.}
\label{tab:asr_st_res}
\end{table}

\subsection{Validation of SPES Hyperparameters}
\label{app:spes_validation}

In this section, we describe the SPES hyperparameter validation process (see \S \ref{sec:experimental_setting}). Due to the high number and wide range of possible hyperparameter values, a full systematic evaluation was computationally prohibitive. Thus, we began with a manual inspection of explanations on a subset of MuST-C dev set samples, narrowing down value ranges for each hyperparameter and setting the perturbation iterations $N^X$ and $N^Y$ at \num{20000} and \num{2000}, respectively, to balance explanation quality and computational cost.

Next, we conducted a systematic validation to select the best configuration using the \textit{deletion} metric (see \S \ref{sec:evaluation}). 
We validated spectrogram perturbation hyperparameters, except for the threshold length $\tau$, in \ref{app:spectrogram_validation} and output token perturbation hyperparameters in  \ref{app:token_validation}.
Given the reduced, yet still numerous, hyperparameter combinations after initial filtering, this validation used a subsample of $130$ utterances, randomly chosen between $5$ and $10$ seconds to represent typical lengths and avoid rare length biases. Finally, we validated $\tau$ on the full validation set for both ASR and ST tasks to examine behavior across diverse lengths.

\begin{table}[b!]
\centering
\small
\begin{tabular}{c|c|c||c|c}
\toprule
\multirow{2}{*}{\textbf{$k$}} & \multirow{2}{*}{\textbf{$\sigma$}} & \multirow{2}{*}{$\mathbf{p_{spec}}$} & \multicolumn{2}{c}{\textbf{Deletion ($\uparrow$)}} \\
\cline{4-5}
 & & & \textbf{w/o norm} & \textbf{w/ norm} \\
\hline
\hline
\multirow{6}{*}{\makecell{500 \\ 1000 \\ 2000}} &  \multirow{3}{*}{0}                         & 0.3                   & 91.75 & 91.90 \\
                                  &                           & 0.5                   & 91.95 & 92.87 \\
                                  &                           & 0.7                   & 84.90 & 91.94 \\
\cline{2-5}
                                   & \multirow{3}{*}{1}        & 0.3                   & 91.92 & 91.88 \\
                                  &                           & 0.5                   & 92.29 & 93.32 \\
                                   &                           & 0.7                   & 85.26 & 91.94 \\
\hline
\multirow{6}{*}{\makecell{1000 \\ 1500 \\ 2000}} & \multirow{3}{*}{0}        & 0.3                   & 93.38 & 93.09 \\
                                    &                           & 0.5                   & 93.41 & 94.56 \\
                                    &                           & 0.7                   & 84.49 & 92.45 \\
\cline{2-5}
                                    & \multirow{3}{*}{1}        & 0.3                   & 92.93 & 92.62 \\
                                    &                           & 0.5                   & 93.71 & 94.71 \\
                                    &                           & 0.7                   & 84.82 & 92.65 \\
\hline
\multirow{6}{*}{\makecell{\textbf{2000} \\ \textbf{2500} \\ \textbf{3000}}} & \multirow{3}{*}{\textbf{0}}        & 0.3              
     & 93.73 & 93.28 \\
                                    &                           & \textbf{0.5}                   & 94.12 & \cellcolor{gray!30}\textbf{95.37} \\
                                    &                           & 0.7                   & 83.02 & 91.88 \\
\cline{2-5}
                                    & \multirow{3}{*}{1}        & 0.3                   & 93.62 & 93.18 \\
                                    &                           & 0.5                   & 93.78 & 95.05 \\
                                    &                           & 0.7                   & 82.45 & 91.93 \\
\hline
\multirow{6}{*}{\makecell{3000 \\ 3500 \\ 4000}} & \multirow{3}{*}{0}        & 0.3                   & 92.99 & 92.51 \\
                                    &                           & 0.5                   & 92.39 & 94.24 \\
                                    &                           & 0.7                   & 80.01 & 90.25 \\
 \cline{2-5}
                                    & \multirow{3}{*}{1}        & 0.3                   & 92.37 & 91.75 \\
                                    &                           & 0.5                   & 91.35 & 93.94 \\
                                    &                           & 0.7                   & 78.55 & 88.70 \\
\hline
\multirow{6}{*}{\makecell{1000 \\ 2500 \\ 5000}} & \multirow{3}{*}{0}        & 0.3                   & 92.56 & 92.07 \\
                                    &                           & 0.5                   & 92.46 & 93.94 \\
                                    &                           & 0.7                   & 83.00 & 91.06 \\
 \cline{2-5}
                                    & \multirow{3}{*}{1}        & 0.3                   & 91.61 & 91.17 \\
                                    &                           & 0.5                   & 92.29 & 93.87 \\
                                    &                           & 0.7                   & 82.87 & 91.06 \\
\hline
\multicolumn{3}{c||}{Avg.} & 89.47 & \textbf{92.04} \\
\bottomrule
\end{tabular}
\caption{Deletion scores to validate $k$, $\sigma$, $p_{spec}$.}
\label{tab:validation_spec}
\end{table}

\subsubsection{Perturbing Spectrograms}
\label{app:spectrogram_validation}

For the perturbation of spectrograms (see \S \ref{sec:spectrogram_method}), we validate three parameters: 
\textit{i)} the number of patches per second ($\phi$), which determines the granularity of the perturbation; 
\textit{ii)} the sigma value ($\sigma$) in SLIC, which controls the smoothness of segmentation contours;
\textit{iii)} the perturbation probability ($p_{spec}$), which defines the probability of each patch being perturbed at each iteration. At this stage, we also validated the effectiveness of applying standard normalization to the token-level saliency maps before aggregating them into a sentence-level explanation. Table \ref{tab:validation_spec} presents the deletion scores for all these aspects.

\paragraph{Number of Patches per Second ($\phi$)} 
As described in \S \ref{sec:methodology}, our segmentation of the spectrogram adopts a multi-scale approach, using three levels of granularity achieved using three $\phi$ values. As our validation set is quite homogeneous in terms of duration (in the $[5, 10]$ seconds range) and the validation procedure is computationally expensive, we do not adjust the number of patches based on duration and optimize three values for the numbers of clusters \textit{k}. Then, we derive $\phi$ by dividing the best \textit{k} values by $5$ seconds. This choice aims to maximize the performance of the approach with fixed number of patches in our ablation study (\textit{- duration adaptation}, see \S\ref{sec:ablation}), ensuring that our method is not advantaged by more favourable hyperparameters. Among the various configuration tested, the combination $k=[2000, 2500, 3000]$ consistently yields better results with the only exception of the results obtained with $p_{spec} = 0.7$. Accordingly, our experiments with SPES use \num{400}, \num{500}, and \num{600} patches per second as the three clustering scales.

\paragraph{Sigma ($\sigma$)}
The preliminary manual inspection revealed that increasing $\sigma$ above $1$ results in patch shapes similar to grids. Therefore, we considered two values for $\sigma$: $0$ and $1$. The former, the default value in the SLIC implementation in scikit-image, provides sharper patch contours, while the latter smooths them. The best value for $\sigma$ varies across the different configurations, with no clear trend emerging. This indicates that both values are effective in terms of perturbation.  However, for the selected number of patches ($k=[2000, 2500, 3000]$), $\sigma=0$ yields better results. For this reason, we set $\sigma$ to 0 in our experiments.

\paragraph{Perturbation Probability ($p_{spec}$)}
The initial manual inspection revealed that high and low $p_{spec}$ values do not produce good explanations. For this reason, we tested three $p_{spec}$ values: $0.3$, $0.5$, and $0.7$. In all configurations of the other hyperparameters, the intermediate value provided better results. This suggests that a $p_{spec}$ of $0.3$ may be too low, as the model remains robust with only a few perturbed features, while a value of $0.7$ may result in excessive information loss. In light of this, a $p_{spec}$ of $0.5$ proves to be an effective compromise.

\paragraph{Normalization} 
For each parameter combination, we computed the scores using token-wise standard normalization. As discussed in \S \ref{sec:spes_perturbation_impact}, normalization levels off the differences between the score distributions of different tokens, which depend on the original probability distributions. In most combinations (specifically $20$ out of the $30$ considered), including that with the highest score ($k=[2000, 2500, 3000]$, $\sigma=0$, $p_{spec}=0.5$), normalization turns out to be beneficial, and, on average, the scores with normalization were $2.57$ points higher than those without normalizaton. Therefore, when aggregating token-level explanations at the sentence level, we use token-level normalization by default. 

In summary, the selected parameters for spectrogram perturbation are: $\phi=[400, 500, 600]$, $\sigma=0$, and $p_{spec}=0.5$. Our validation also proved that normalizing the token-level saliency maps leads to better sentence-level explanations.

\begin{table}[tb!]
\centering
\small
\begin{tabular}{c||c|c}
\toprule
\multirow{2}{*}{$\mathbf{p_{tok}}$} & \multicolumn{2}{c}{\textbf{Deletion ($\uparrow$)}} \\
\cline{2-3}
 & \textbf{Joint} & \textbf{Separate} \\
\hline
\hline
0.1                   & 83.72 & 94.49                      \\
0.2                   & 79.77 & 94.48                      \\
0.3                   & 77.48 & 94.45                      \\
\textbf{0.4}                   & 74.36 & \cellcolor{gray!30}\textbf{94.50}                    \\
0.5                   & 71.02 & 94.49                      \\
0.6                   & 69.79 & 94.48                      \\
0.7                   & 66.57 & 94.47                      \\
0.8                   & 63.31 & 94.42                      \\
0.9                   & 62.71 & 94.42       \\
\hline
 \multicolumn{1}{c||}{Avg.} & 72.08 & \textbf{94.47} \\
\bottomrule
\end{tabular}
\caption{Deletion scores to validate $p_{tok}$ in SPES.}
\label{tab:p_tok}
\end{table}

\subsubsection{Perturbing Previous Output Tokens}
\label{app:token_validation}

After determining an effective combination of parameters for the spectrogram perturbation, we moved to validate the perturbation of the previous output tokens (see \S \ref{subsec:token_method}).
Specifically, we analyzed two aspects: \textit{i)} whether it is better to jointly perturb the speech input and previous output tokens or not, and \textit{ii)} determining the optimal value of the perturbation probability $p_{tok}$.

\paragraph{Joint vs Separate Perturbation}
Previous output tokens can be perturbed in two ways: jointly with the spectrogram or separately. In the joint approach, both the spectrogram features and the previous output tokens are perturbed simultaneously, potentially revealing interdependencies between them for next-token predictions, and saliency scores are accordingly assigned to the perturbed elements of both sides. However, since the previous tokens are fewer than the speech features, they are perturbed more frequently, which can ``blur'' explanations on the spectrogram. Table \ref{tab:p_tok} presents the deletion scores for both joint and separate perturbations. The results consistently show higher scores for separate perturbations, indicating that joint perturbation degrades performance. Consequently, in our experiments, we compute saliency scores separately for the speech input and the previous output tokens in SPES.

\paragraph{Perturbation Probability ($p_{tok})$} We considered $p_{tok}$ values in the range $[0.1, 0.9]$. 
The differences in scores were slight (see Table \ref{tab:p_tok}). 
However, similar to the trend observed for $p_{spec}$, performance degradation was noted with values greater than $0.7$, indicating that excessive information removal hinders effective perturbations. Unlike $p_{spec}$, lower values for $p_{tok}$ were effective, most likely due to the lower cardinality and lower redundancy of information of the previous tokens compared to the features in the spectrograms. The best score was achieved with $p_{tok}$ set to $0.4$.

In summary, in all our experiments the previous output tokens are perturbed separately from the spectrogram, with $p_{tok}$ equal to $0.4$.

\begin{figure}[tb!]
    \centering
    \begin{subfigure}[b]{0.48\textwidth}
    \includegraphics[width=\textwidth]{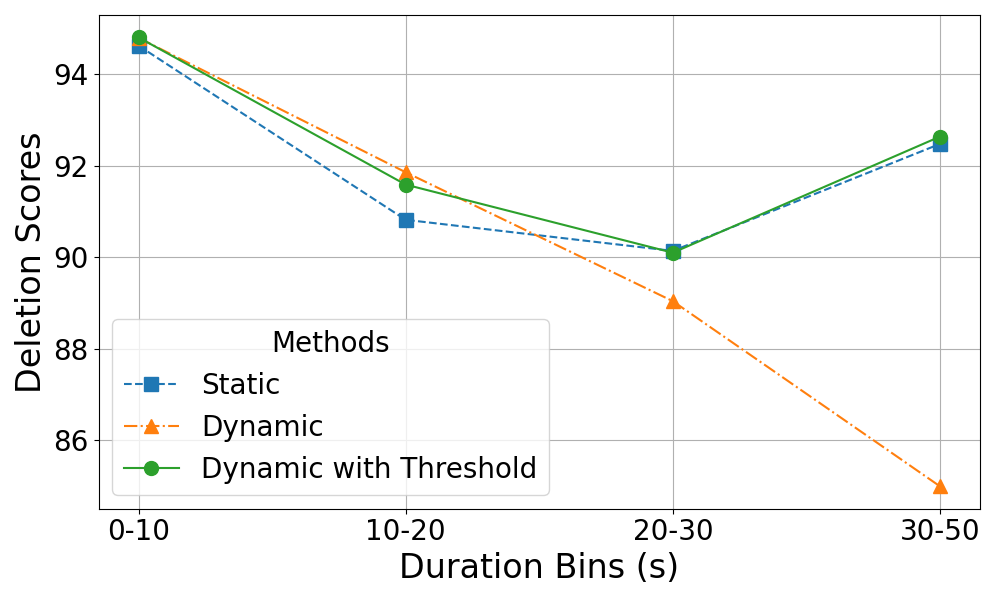}
    \end{subfigure}
    \caption{Variation of Deletion scores across samples of different lengths.}
    \label{fig:duration}
\end{figure}

\subsubsection{Validation of the Dynamic Increase of Number of Patches and $\tau$}
\label{app:tau}

As discussed in \S \ref{sec:spectrogram_method}, dynamically increasing the number of patches for long audio samples significantly raises the number of patch combinations, potentially reducing explanation quality. Fig. \ref{fig:duration} shows deletion scores in ASR across input durations on the validation set, comparing a dynamic patch increase based on duration (\textit{Dynamic}) versus a fixed number of patches (\textit{Static}) with \textit{k} set to $[\num{2000}, \num{2500}, \num{3000}]$. While \textit{Dynamic} performs better for samples up to $20$ seconds, its deletion score declines linearly with duration. In contrast, \textit{Static} remains stable scoring $7.49$ points higher than \textit{Dynamic} for samples over $30$ seconds.

\begin{table}[t!]
\centering
\small
\begin{tabular}{c||c|c|c}
\toprule
\multirow{2}{*}{\textbf{$\tau$}} & \multicolumn{3}{c}{\textbf{Deletion}} \\
\cline{2-4}
& \textbf{en (\(\uparrow\))} & \textbf{en$\rightarrow$de (\(\downarrow\))} & \textbf{en$\rightarrow$es (\(\downarrow\))} \\
\hline
\hline
2.5                     & 93.01 & 2.47 & 2.91 \\ 
\textbf{5.0}                     & 93.91 & \cellcolor{gray!30}\textbf{2.34} & \cellcolor{gray!30}\textbf{2.58} \\ 
\textbf{7.5} & \cellcolor{gray!30}\textbf{94.03} & 2.41 & 2.62 \\
10.0                    & 93.91 & 2.49 & 2.77 \\
\bottomrule
\end{tabular}
\caption{Deletion scores to validate the duration threshold $\tau$ (in seconds) for stopping the linear increase in the number of patches.}
\label{tab:duration}
\end{table}

To mitigate this issue, we introduced the threshold hyperparameter $\tau$ (see \S \ref{sec:spectrogram_method}), which limits dynamic increases in the number of patches beyond certain audio lengths. Table \ref{tab:duration} presents the validation results for $\tau$ on the full ASR and ST validation sets. For ASR, the optimal threshold is set at $7.5$ seconds, while for ST, a lower threshold of $5.0$ seconds is established. In Fig. \ref{fig:duration}, this \textit{Dynamic with Threshold} approach effectively avoids the decline in faithfulness observed with \textit{Dynamic}, consistently outperforming \textit{Static} across all sample lengths, including those exceeding $30$ seconds.

In summary, ASR uses $k$ = $[3,000, 3,750, 4,500]$ for audio longer than $7.5$ seconds, while ST uses $k$ = $[2,000, 2,500, 3,000]$ for audio over $5.0$ seconds.

\begin{table}[t!]
\centering
\small
\begin{tabular}{c||c}
\toprule
\multicolumn{1}{l||}{$\mathbf{p_{spec}}$} & \multicolumn{1}{l}{\textbf{Deletion ($\uparrow$)}} \\
\hline
\hline
0.5                   & 74.79                      \\
0.6                   & 76.63                      \\
\textbf{0.7}                   & \cellcolor{gray!30}\textbf{79.29}                      \\
0.8                   & 78.49                      \\
0.9                   & 58.67       \\
\bottomrule
\end{tabular}
\caption{Deletion scores to validate $p_{spec}$ in the \textit{feature-wise} configuration.}
\label{tab:validation_pixel_wise_spec}
\end{table}

\begin{table}[tb!]
\centering
\small
\begin{tabular}{c||c}
\toprule
\multicolumn{1}{l||}{$\mathbf{p_{tok}}$} & \multicolumn{1}{l}{\textbf{Deletion ($\uparrow$)}} \\
\hline
\hline
\textbf{0.1}                   & \cellcolor{gray!30}\textbf{77.38}                      \\
0.2                   & 77.31                      \\
0.3                   & 77.29                      \\
0.4          & 77.21                    \\
0.5                   & 77.17                      \\
0.6                   & 77.18                     \\
0.7                   & 77.21                      \\
0.8                   & 77.10                      \\
0.9                   & 77.06       \\
\bottomrule
\end{tabular}
\caption{Deletion scores to validate $p_{tok}$ in the \textit{feature-wise} configuration.}
\label{tab:validation_pixel_wise_tok}
\end{table}

\subsection{Validation of Baselines Hyperparameters}

To ensure a fair comparison with SPES, we conducted a validation procedure to determine the optimal hyperparameters for the \textit{feature-wise} and \textit{Bubble Noise} baselines discussed in \S \ref{sec:experimental_setting}.

\subsubsection{Feature-wise Perturbation}
\label{app:features_wise_validation}

Since no segmentation of the spectrogram is involved in \textit{feature-wise}, the only parameters to be validated are $p_{spec}$ and $p_{tok}$.
The preliminary manual inspection revealed that low values for $p_{spec}$ were not effective in providing high deletion scores. Since the perturbation involves only sparse single features rather than groups of contiguous features (i.e., the patches), it is reasonable that the model is robust in recovering information even when a large percentage of features is perturbed. Therefore, we focused on values from $0.5$ onward, whose deletion scores are shown in Table \ref{tab:validation_pixel_wise_spec}. The best result was achieved with $0.7$.

Table \ref{tab:validation_pixel_wise_tok} shows the deletion scores for $p_{tok}$, with the perturbation of the previous output tokens carried out separately, as in SPES. Similar to $p_{tok}$ in SPES, the differences among the scores are low, with lower values generally providing higher scores. However, in this case, $p_{tok}$ set to $0.1$ yielded the highest result.

In summary, for the experiments with the feature-wise baseline reported in \S \ref{sec:results}, we set $p_{spec}$ to $0.7$ and $p_{tok}$ to $0.1$.

\subsubsection{Bubble Noise Perturbation}
\label{app:bubble}

We implement the Bubble Noise perturbation method \citep{mandel13_bubble, trinh20_interspeech, trinh21_bubble}, enhancing it to also produce explanations for previous output tokens (see \S \ref{sec:experimental_setting}).
To perturb the spectrogram, we generate random noise that matches the intensity range of the speech input.\footnote{We also tested white noise at various levels of SNR relative to the input spectrogram. However, noise matching the intensity range of the input spectrogram performed better.}
This noise replaces the spectrogram everywhere except within randomly placed ellipsoidal bubbles, where it is attenuated and then added back to the spectrogram.

Regarding bubble size and count, previous works use various numbers of bubbles per second, ranging from $18$ to $80$, all with a fixed size of $0.35$ seconds and a height of approximately $180$ Hz. We explore different configurations within a range that encompasses the values tested in earlier studies. Specifically, we aim to cover about half of the spectrogram with noise, aligning with SPES’s perturbation level ($p_{spec}=0.5$). Table \ref{tab:bubble} presents deletion scores for three levels of bubble granularity tested with $N^X$ = \num{1000}. The configuration with $10$ bubbles per second yielded the best results, achieving deletion scores that were $10.33$ and $12.43$ points higher than the configurations with $35$ and $115$ bubbles per second, respectively.

In summary, Bubble Noise perturbation was set to $10$ bubbles per second, each covering $\sim$$0.43$ seconds in width and $\sim$$31$ mel bins in height.

\begin{table}[tb!]
\centering
\small
\begin{tabular}{c|c|c||c}
\toprule
\textbf{Bub/s} & \textbf{Width (s)} & \textbf{Height (mel)} & \textbf{Deletion ($\uparrow$)} \\ 
\hline
\hline
\cellcolor{gray!30}\textbf{10} & \cellcolor{gray!30}$\sim$$0.43$ & \cellcolor{gray!30}$\sim$$31$ & \cellcolor{gray!30}\textbf{69.40} \\ 
35 & $\sim$$0.25$ & $\sim$$15$ & 59.07                      \\ 
115 & $\sim$$0.13$ & $\sim$$9$ & 56.97                      \\ 
\bottomrule
\end{tabular}
\caption{Deletion scores to validate bubble size and count.}
\label{tab:bubble}
\end{table}

\subsection{Examples of Explanations}
\label{app:examples}

\begin{figure}[!ht]
    \centering
    \begin{subfigure}[b]{0.455\textwidth}
        \centering
        \includegraphics[width=\textwidth]{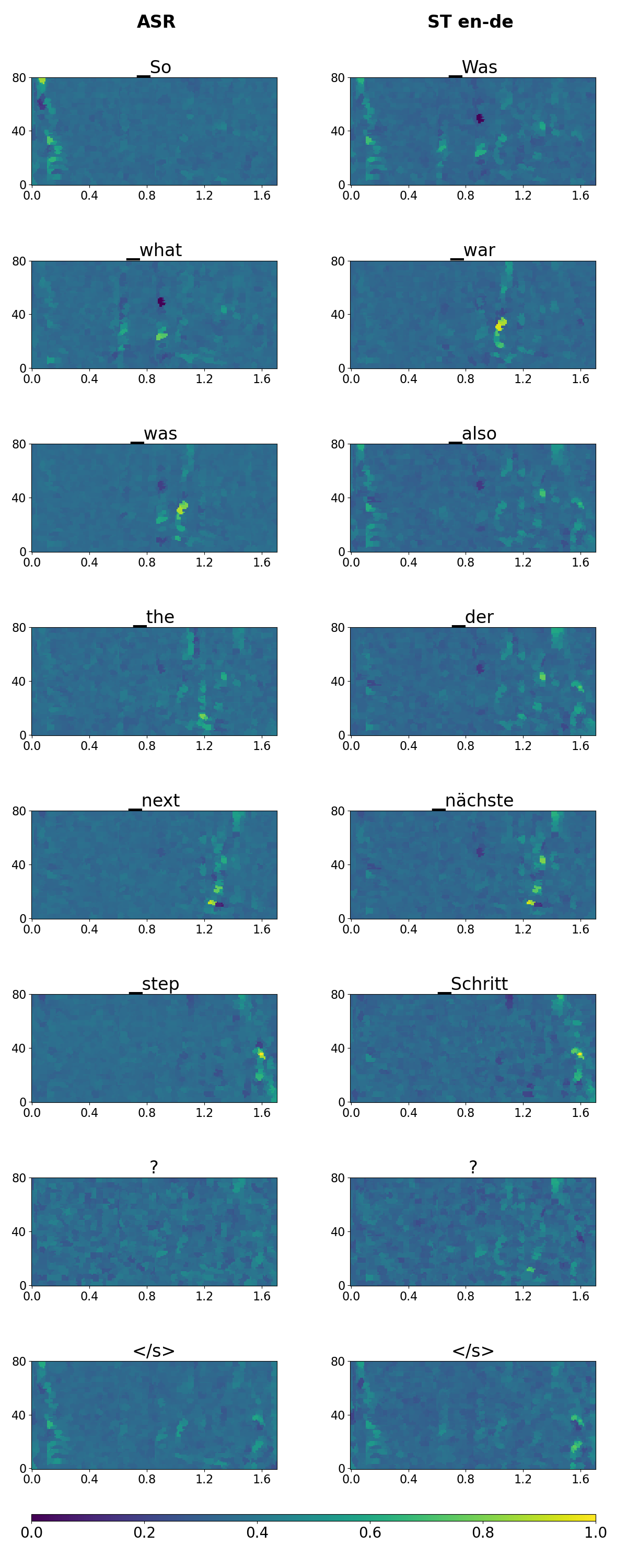}
    \end{subfigure}
    \vspace{-1em}
    \begin{subfigure}[b]{0.44\textwidth}
        \centering
        \includegraphics[width=\textwidth]{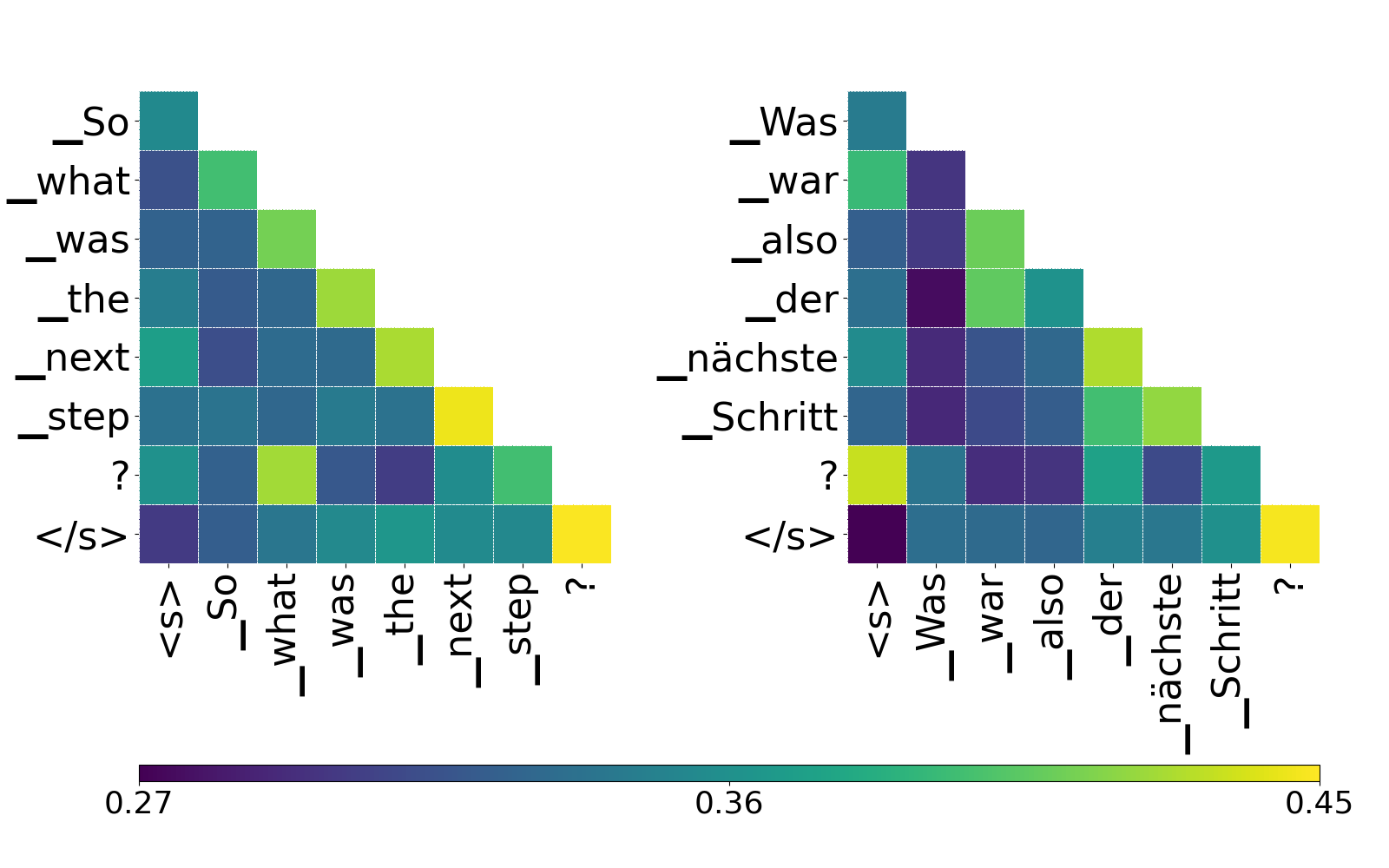}
    \end{subfigure}
    \caption{$S^{X}$ (top) and $S^{Y^{(k-1)}}$ (bottom) for ASR and ST on the utterance ``So what was the next step?''. In $S^{Y^{(k-1)}}$, rows show the tokens being explained, and columns 
    the previous output tokens.}
    \label{fig:examples}
\end{figure}

Fig. \ref{fig:examples} shows $S^{X}$ and $S^{Y^{(k-1)}}$ for the transcription and the translation into German of the utterance ``So what was the next step?''.
These visualizations highlight the following points, also detailed in \S \ref{sec:plausibility}.

\begin{figure}[H]
   \centering
   \includegraphics[width=0.43\textwidth]{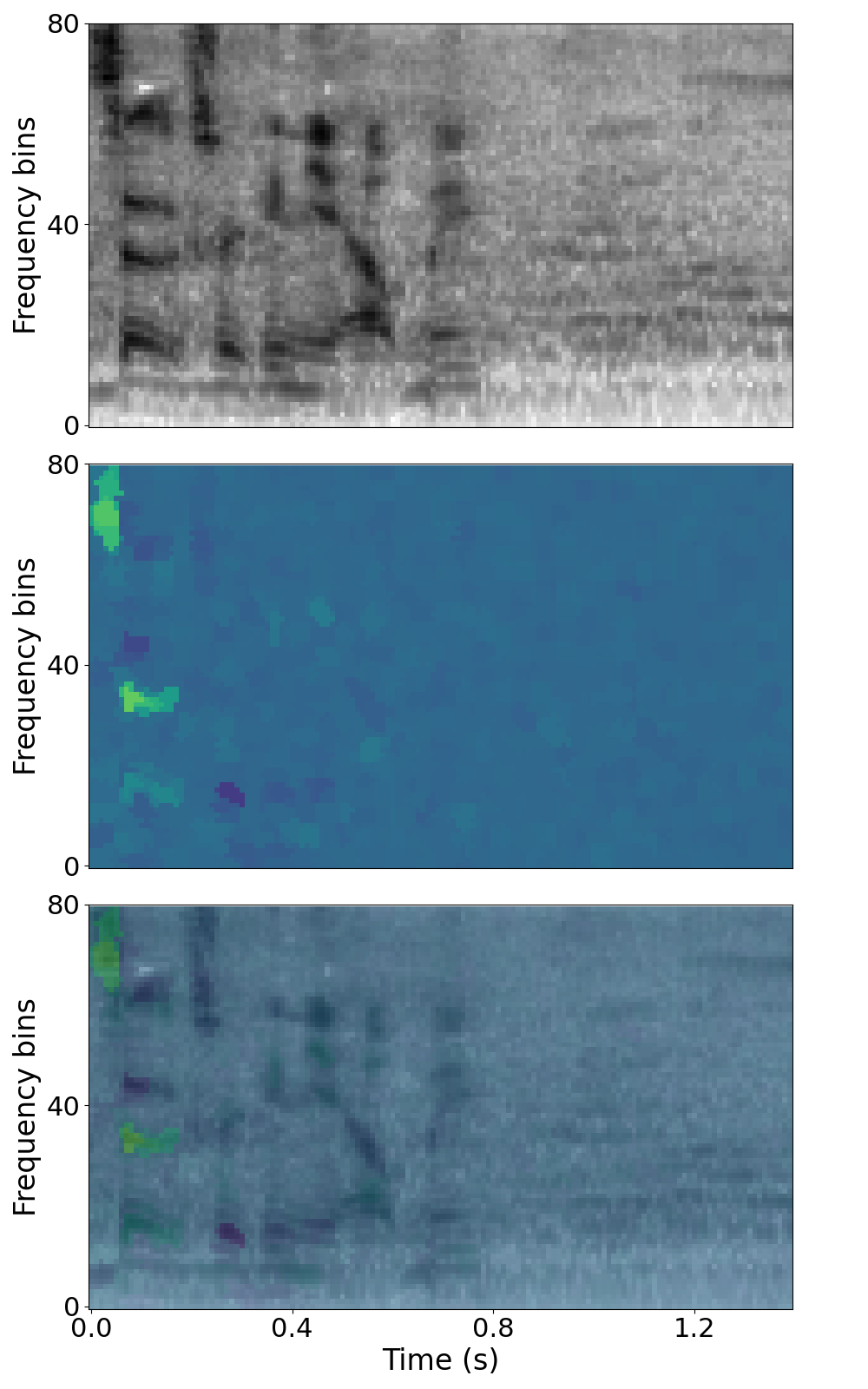}
    \caption{Example of saliency map (middle) for the token \texttt{so} (ASR), along with the corresponding spectrogram (top) and the map overlayed on the spectrogram (bottom).}
    \label{fig:so}
\end{figure}

\begin{enumerate}
    \item Comparing ASR and ST $S^{X}$ maps reveals similar patterns; for example, the German tokens \texttt{war}, \texttt{n\"achste}, and \texttt{Schritt} align with the English counterparts \texttt{was}, \texttt{next}, and \texttt{step}, indicating a reliance on similar phonetic sequences.

    \item In ASR $S^{X}$, salient areas move forward with each token, showing a monotonic progression in time (see \S \ref{sec:plausibility_spectrogram}). 
    This effect is less pronounced in ST due to differences in word order, such as German \texttt{Was} and \texttt{also} (en. \textit{what} and \textit{so}) reversing the English order.

    \item Most $S^{X}$ maps are frequency-localized. For instance, \texttt{So} (ASR) shows high saliency in the high-frequency range typical of alveolar sibilants and mid-frequency ranges for vowel formants (see \S \ref{sec:plausibility_spectrogram} and Fig. \ref{fig:so}).
    
    \item Some $S^{X}$ maps lack clear salient areas, such as those for \texttt{?} in both ASR and ST (see \S \ref{subsec:spectogram-output}).

    \item In $S^{Y^{(k-1)}}$, the first and last tokens generally exhibit high saliency, with some intermediate tokens showing similar effects (see \S \ref{subsec:analyses_tokens}).
    For example, in ASR, \texttt{?} is influenced by \texttt{what}, and in ST, \texttt{der} (en. \textit{the}) by \texttt{war} (en. \textit{was}).
\end{enumerate}

\clearpage

\section{Complementary Results}
\label{app:complementary-results}

\begin{figure}[H]
    \centering
    \begin{subfigure}[b]{0.54\textwidth}
    \includegraphics[width=\textwidth]{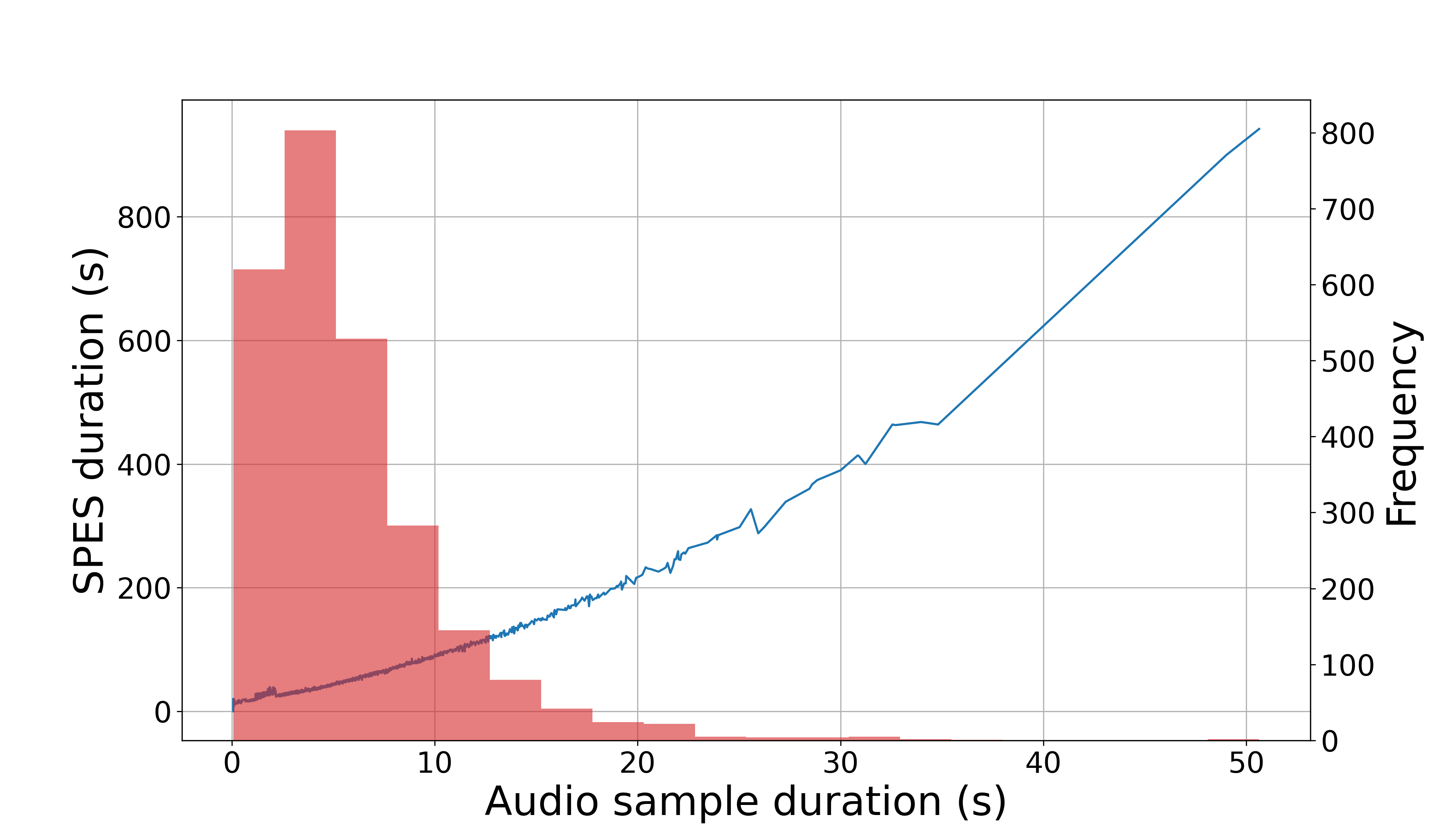}
    \end{subfigure}
    \caption{SPES explanation generation time across varying sample lengths and distribution of samples by length.}
    \label{fig:time}
\end{figure}

\begin{table}[H]
\small
\centering
\begin{tabular}{l|cc}
\toprule
 & \multicolumn{2}{c}{\textbf{ASR (en)}} \\
\cline{2-3}
 & \textbf{Deletion} ($\uparrow$) & \textbf{Size} ($\downarrow$) \\
\hline
Bubble Noise & 65.02 & 36.92 \\
SPES fine-grained & 75.22 & 39.54 \\
SPES coarse-grained & \textbf{79.71} & \textbf{26.91} \\
\hline
 & \multicolumn{2}{c}{\textbf{ST (en$\rightarrow$de)}} \\
\cline{2-3}
 & \textbf{Deletion} ($\downarrow$) & \textbf{Size} ($\downarrow$) \\
\hline
Bubble Noise & 10.85 & 43.89 \\
SPES fine-grained & 6.84 & 40.06  \\
SPES coarse-grained & \textbf{5.14} & \textbf{28.58}  \\
\hline
 & \multicolumn{2}{c}{\textbf{ST (en$\rightarrow$es)}} \\
\cline{2-3}
 & \textbf{Deletion} ($\downarrow$) & \textbf{Size} ($\downarrow$) \\
\hline
Bubble Noise & 13.01 & 43.04 \\
SPES fine-grained & 7.41 & 39.66 \\
SPES coarse-grained & \textbf{6.10} & \textbf{27.17}  \\
\bottomrule
\end{tabular}
\caption{Results comparing Bubble Noise and SPES with $N^X$ = $1,000$. SPES is evaluated in both a \textit{fine-grained} setting ($\phi$ validated in Appendix \ref{app:spectrogram_validation}) and a \textit{coarse-grained} setting ($\phi$ = $[20, 40, 60]$), maintaining approximately $10$ unperturbed patches per second---similar to Bubble Noise, which attenuates noise in $10$ bubbles per second.}
\label{tab:results_bubble}
\end{table}

\begin{figure}[H]
    \centering
    \begin{subfigure}[b]{0.45\textwidth}
    \includegraphics[width=\textwidth]{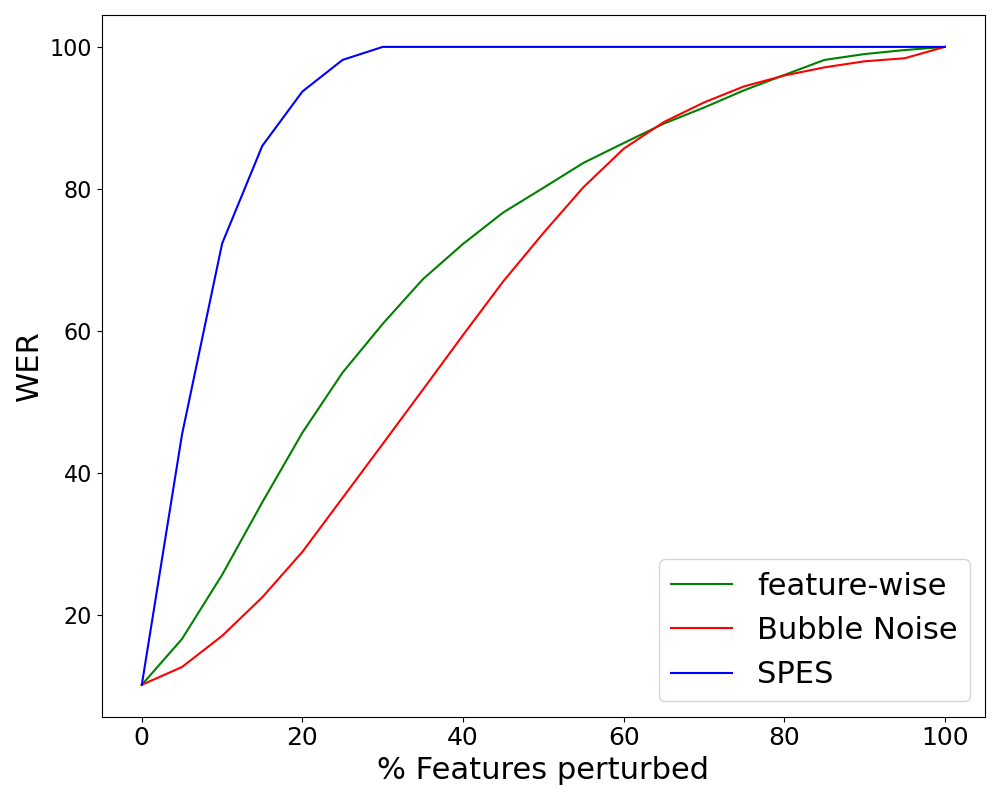}
    \subcaption{ASR (en)}
    \end{subfigure}
    \begin{subfigure}[b]{0.45\textwidth}
    \includegraphics[width=\textwidth]{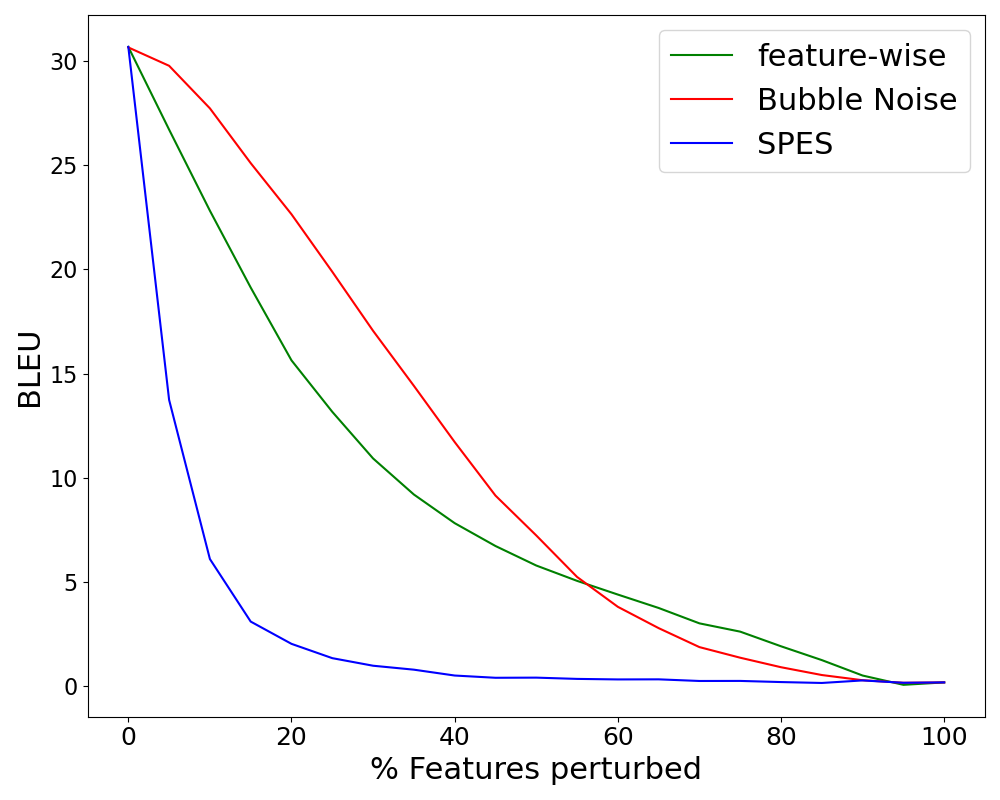}
    \subcaption{ST (en$\rightarrow$de)}
    \end{subfigure}
    \begin{subfigure}[b]{0.45\textwidth}
    \includegraphics[width=\textwidth]{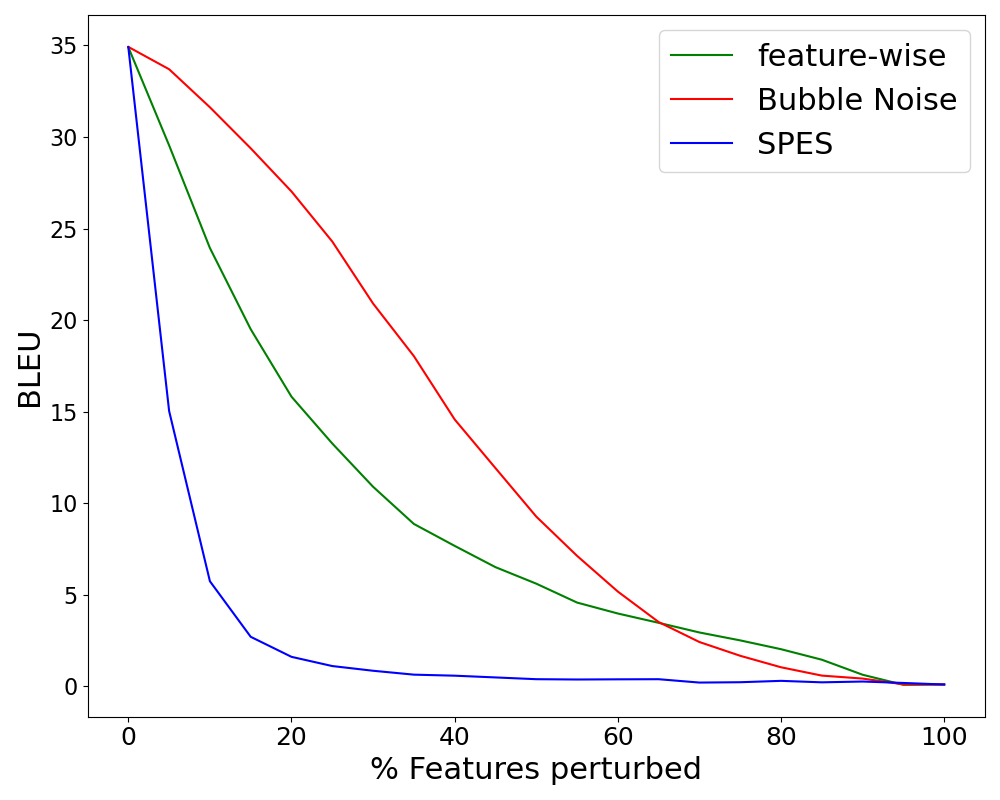}
    \subcaption{ST (en$\rightarrow$es)}
    \end{subfigure}
    \caption{Deletion curves for the \textit{feature-wise} strategy, the \textit{Bubble Noise} technique, and SPES in the ASR and ST tasks.}
    \label{fig:deletion}
\end{figure}

\begin{figure}[H]
    \centering
    \begin{subfigure}[b]{0.45\textwidth}
    \includegraphics[width=\textwidth]{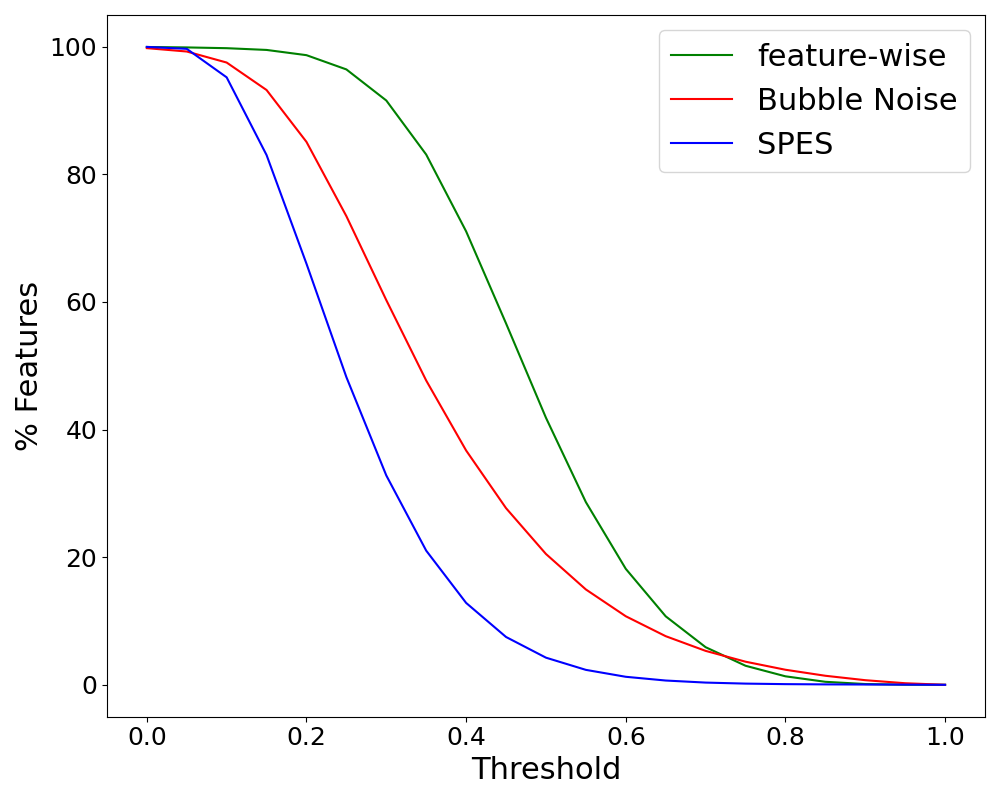}
    \subcaption{ASR (en)}
    \end{subfigure}
    \begin{subfigure}[b]{0.45\textwidth}
    \includegraphics[width=\textwidth]{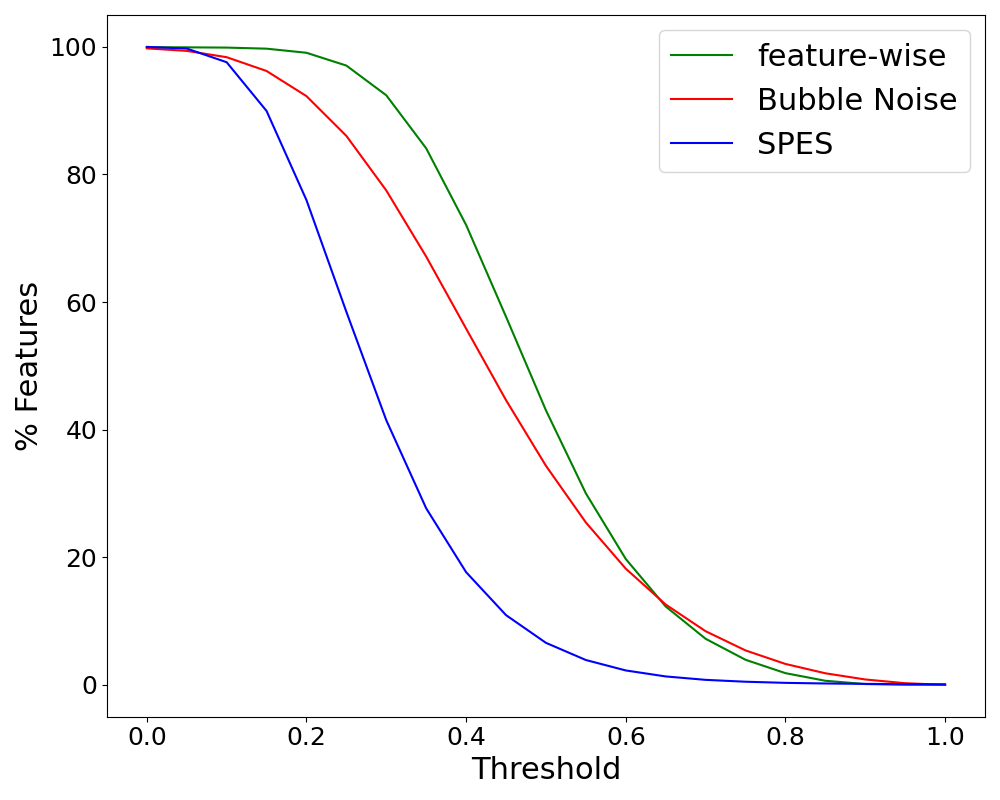}
    \subcaption{ST (en$\rightarrow$de)}
    \end{subfigure}
    \begin{subfigure}[b]{0.45\textwidth}
    \includegraphics[width=\textwidth]{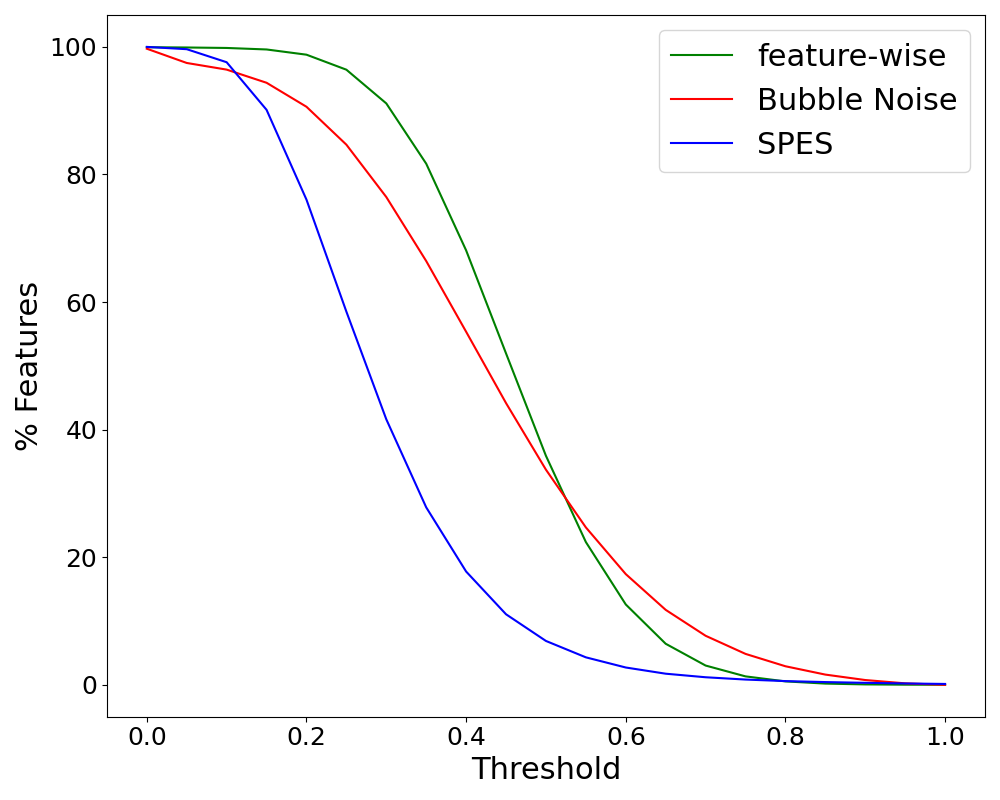}
    \subcaption{ST (en$\rightarrow$es)}
    \end{subfigure}
    \caption{Size curves for the \textit{feature-wise} strategy, the \textit{Bubble Noise} technique, and SPES in the ASR and ST tasks.}
    \label{fig:size}
\end{figure}

\begin{figure}[H]
    \centering
    \begin{subfigure}[b]{0.46\textwidth}
    \includegraphics[width=\textwidth]{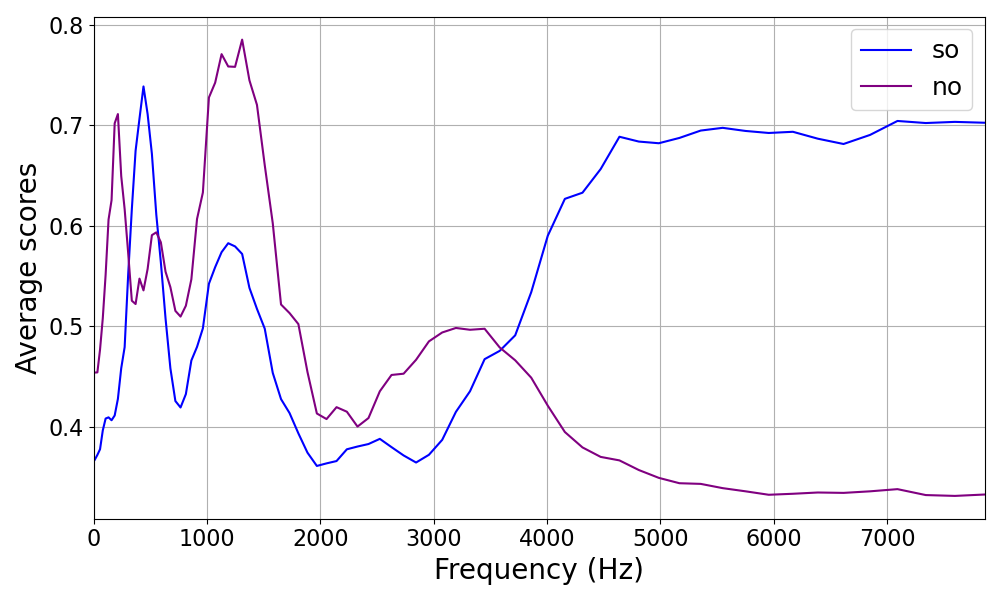}
    \end{subfigure}
    \begin{subfigure}[b]{0.46\textwidth}
    \includegraphics[width=\textwidth]{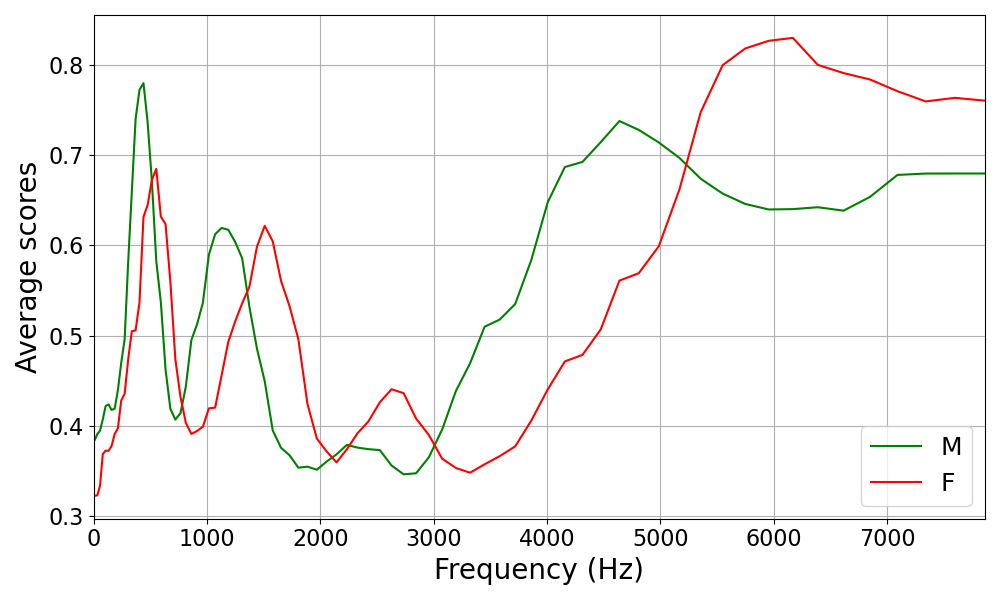}
    \end{subfigure}
    \caption{Average saliency scores over frequency for the minimal pair \textit{so}-\textit{no} (upper figure) and the word \textit{so} uttered by men and women (lower figure).}
    \label{fig:freq}
\end{figure}

\begin{figure}[H]
   \centering
   \includegraphics[width=0.43\textwidth]{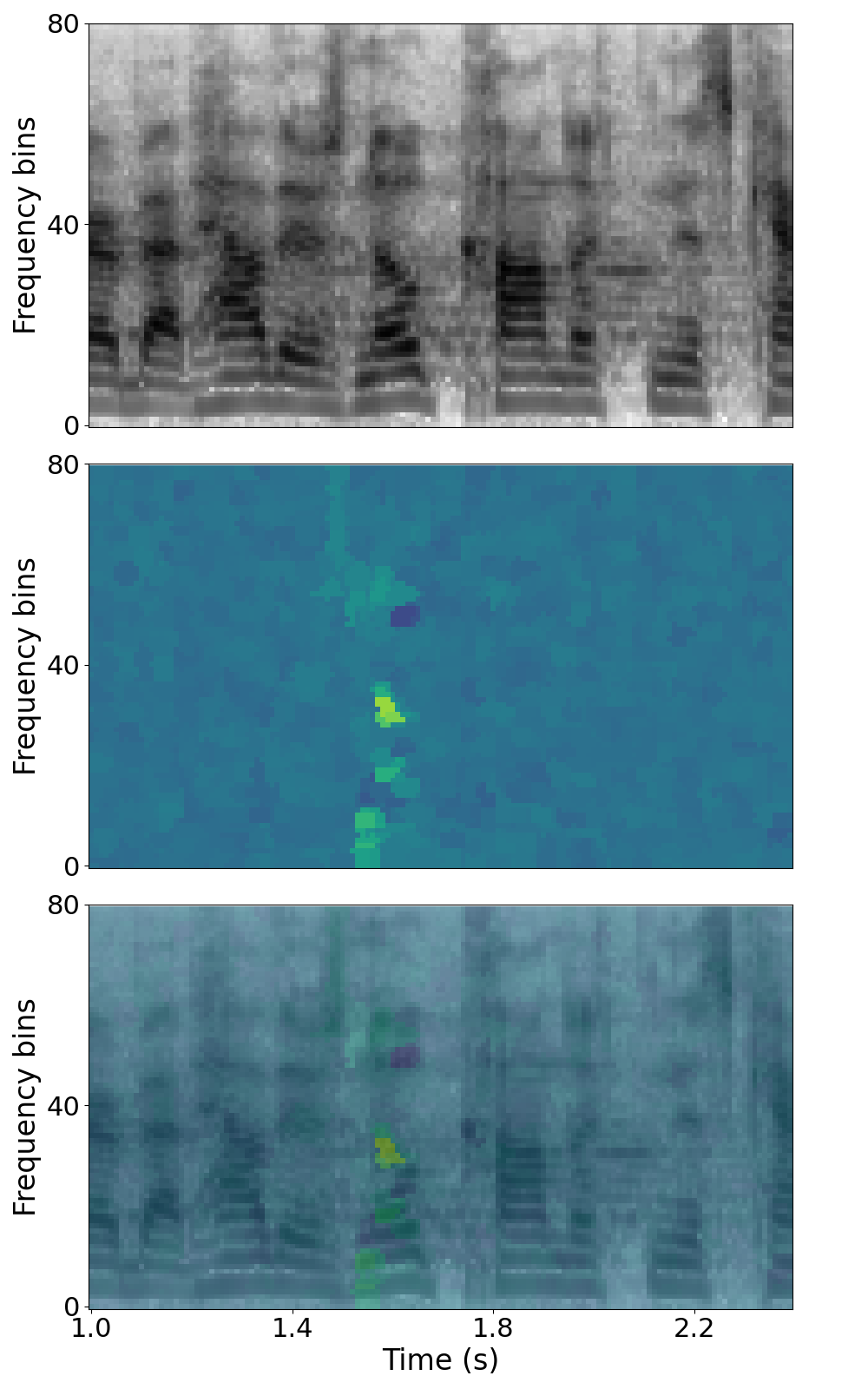}
    \caption{Example of saliency map (middle) for the token \texttt{no} (ASR), along with the corresponding spectrogram (top) and the map overlayed on the spectrogram (bottom).}
    \label{fig:no}
\end{figure}

\begin{figure}[H]
    \centering
    \begin{subfigure}[b]{0.39\textwidth}
    \includegraphics[width=\textwidth]{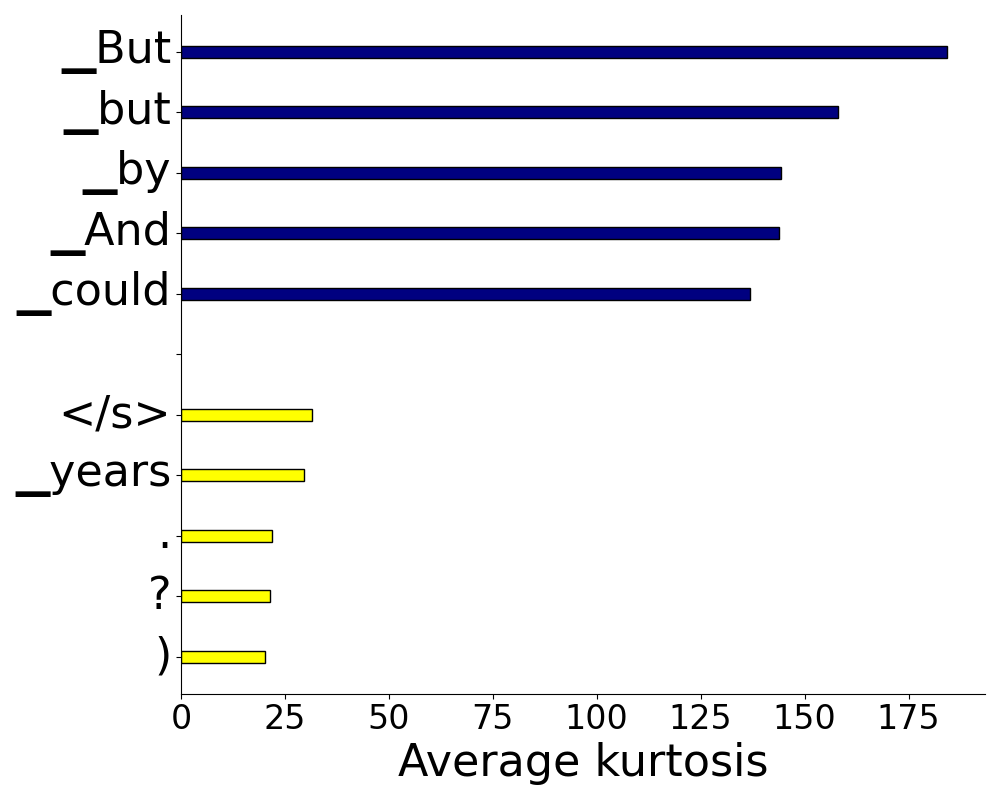}
    \end{subfigure}
    \caption{Top and bottom 5 tokens by kurtosis.}
    \label{fig:kurtosis}
\end{figure}

\begin{figure}[H]
    \centering
    \begin{subfigure}[b]{0.48\textwidth}
    \includegraphics[width=\textwidth]{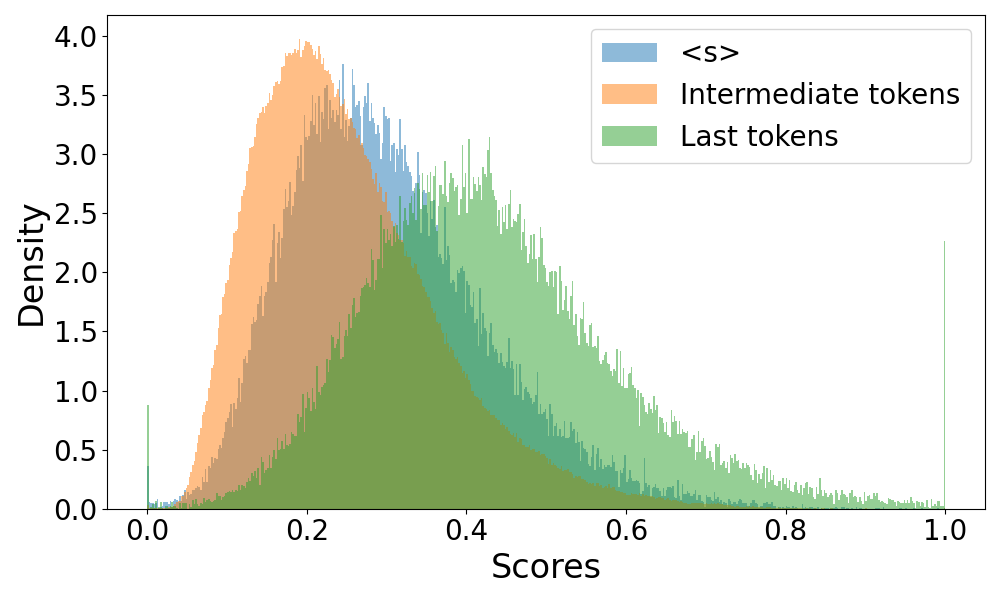}
    \end{subfigure}
    \caption{Distributions of 
    saliency
    scores obtained in ST (en$\rightarrow$de) for three types of tokens: \texttt{<s>}, \textit{intermediate tokens} and tokens preceding the token to be explained (\textit{last tokens}).}
    \label{fig:token_position}
\end{figure}

\begin{table}[H]
\small
\centering
\setlength{\tabcolsep}{3.5pt}
\begin{tabular}{cclc}
\toprule
 Task & Token & \texttt{IT}          & \% \\
 \midrule
\multirow{2}{*}{\textit{ASR en}} & \textbf{)}          & \textbf{(} {[}47{]}                  & 40.5      \\
 & \textbf{?}  & \begin{tabular}[c]{@{}l@{}}\textbf{What} {[}11{]}, \textbf{what} {[}7{]} 
\end{tabular}         & 29.1 \\
\cmidrule{1-4} 
\multirow{2}{*}{ST en$\rightarrow$es}
& \textbf{?}         & \textbf{¿} {[}35{]}, \textbf{"¿} {[}7{]},  
& 82.2      \\
 & \textbf{)}          & \textbf{(} {[}148{]}                 & 33.1      \\
\cmidrule{1-4} 
\multirow{3}{*}{\textit{ST en$\rightarrow$de}} & \textbf{)} & \textbf{(} {[}105{]}
& 87.6 \\
& \textbf{zu} & \begin{tabular}[c]{@{}l@{}}\textbf{um} {[}23{]}, \textbf{zu} {[}11{]}, 
\end{tabular} 
& 29.3 \\
& \textbf{an}         & \textbf{sich} {[}6{]}, \textbf{fängt} {[}4{]} & 25.1     \\
\bottomrule
\end{tabular}
\caption{
Tokens for which the highest saliency score is more often held by an \texttt{IT}. For each token, we provide the overall number of times it is primarily explained by \textit{any} \texttt{IT} (\textit{\%}), and the frequency of each \textit{individual} \texttt{IT} in holding the highest saliency in brackets.}
\label{tab:intermediate_token}
\end{table}

\end{document}